\documentclass[twoside,leqno,twocolumn]{article}

\usepackage[letterpaper]{geometry}

\usepackage{ltexpprt}
\usepackage{hyperref}
\usepackage{algorithm}
\usepackage{algorithmic}
\usepackage{MnSymbol}
\usepackage{subfigure}
\usepackage{graphicx}
\usepackage{mwe}
\usepackage{amsmath}
\usepackage{url}            
\usepackage{booktabs}       
\usepackage{amsfonts}       
\usepackage{nicefrac}       
\usepackage{microtype}      
\usepackage{amsmath}
\usepackage[table]{xcolor}
\usepackage{color}
\usepackage{makecell}
\usepackage{multirow}
\usepackage{color}

\usepackage[numbers,sort&compress]{natbib}
\usepackage[utf8]{inputenc}
\usepackage{soul}
\usepackage{changes}

\usepackage{mathtools}
\begin{document}

\newcommand\relatedversion{}
\renewcommand\relatedversion{\thanks{University of Queensland, \protect\url{{shaofei.shen, chenhao.zhang, alina.bialkowski, miao.xu}@uq.edu.au}. MX is the corresponding author.}}

\title{CaMU: Disentangling Causal Effects in Deep Model Unlearning}
\author{
Shaofei Shen\relatedversion\quad
Chenhao Zhang\footnotemark[1]\quad
Alina Bialkowski\footnotemark[1]\quad
Weitong Chen\thanks{University of Adelaide,  \protect\url{t.chen@adelaide.edu.au}}\quad
Miao Xu\footnotemark[1]
}

\newcommand{\fix}{\marginpar{FIX}}
\newcommand{\new}{\marginpar{NEW}}

\date{}

\maketitle


\fancyfoot[R]{\scriptsize{Copyright \textcopyright\ 2024 by SIAM\\
Unauthorized reproduction of this article is prohibited}}





\begin{abstract} \small\baselineskip=9pt 
Machine unlearning requires removing the information of forgetting data while keeping the necessary information of remaining data. Despite recent advancements in this area, existing methodologies mainly focus on the effect of removing forgetting data without considering the negative impact this can have on the information of the remaining data, resulting in significant performance degradation after data removal. Although some methods try to repair the performance of remaining data after removal, the forgotten information can also return after repair. Such an issue is due to the intricate intertwining of the forgetting and remaining data. Without adequately differentiating the influence of these two kinds of data on the model, existing algorithms take the risk of either inadequate removal of the forgetting data or unnecessary loss of valuable information from the remaining data. To address this shortcoming, the present study undertakes a causal analysis of the unlearning and introduces a novel framework termed Causal Machine Unlearning (CaMU). This framework adds intervention on the information of remaining data to disentangle the causal effects between forgetting data and remaining data. Then CaMU eliminates the causal impact associated with forgetting data while concurrently preserving the causal relevance of the remaining data. Comprehensive empirical results on various datasets and models suggest that CaMU enhances performance on the remaining data and effectively minimizes the influences of forgetting data. Notably, this work is the first to interpret deep model unlearning tasks from a new perspective of causality and provide a solution based on causal analysis, which opens up new possibilities for future research in deep model unlearning. 
\end{abstract}


\section{Introduction}\label{s1}

Machine unlearning is the deliberate process of erasing information that a machine-learning model has previously acquired from training data~\cite{acmsurvey,RememberWhatYouWanttoForget,arxivsurvey}. It plays a crucial role in preserving privacy by allowing the removal of sensitive information~\cite{RememberWhatYouWanttoForget}. With the widespread adoption of deep learning models, the concept of deep model unlearning~\cite{boundary,ts,sisa,unroll} has gained significant attention, particularly in contexts dealing with sensitive data, such as in recommender systems or medical prediction~\cite{RecommendationUnlearning}. The objective of machine unlearning is to selectively remove information associated with \emph{forgetting data} from the pre-unlearning model while retaining the knowledge contained in the \emph{remaining data}. 

Current research in deep model unlearning can be categorized into two main approaches based on the availability of the remaining data. One category, known as \emph{training}-based methods, involves fine-tuning the model using the remaining data~\cite{sisa}. The other category, referred to as \emph{adjustment}-based methods, utilizes the forgetting data only to adjust the model~\cite{unroll, ts, boundary}. Despite the success of these methods in certain tasks, their overall performance remains unsatisfactory due to the potential occurrence of \emph{insufficient unlearning} or \emph{excessive unlearning}~\cite{acmsurvey, arxivsurvey}. 

Despite the fact that insufficient unlearning and excessive unlearning diverge in the extent to which they remove information, they stem from a common underlying cause: the intricate intertwining of information between the forgetting data and remaining data. This intertwining may be attributed to the knowledge shared by both types of data, such as background identification and feature extraction. When these two types of data are highly interdependent, preserving common latent information becomes essential; otherwise, excessive unlearning may occur~\cite{arxivsurvey, Variational}. However, if too much information is retained, the post-unlearning model's performance on forgetting data may experience minimal alteration, resulting in insufficient unlearning~\cite{GKT,boundary}.

\begin{figure}[t]
    \centering
    \subfigure[Conceptual overlap in forgetting and remaining data]{
        \begin{minipage}[htbp]{0.45\linewidth}
            \centering        
            \includegraphics[width=\textwidth,height=\textwidth]{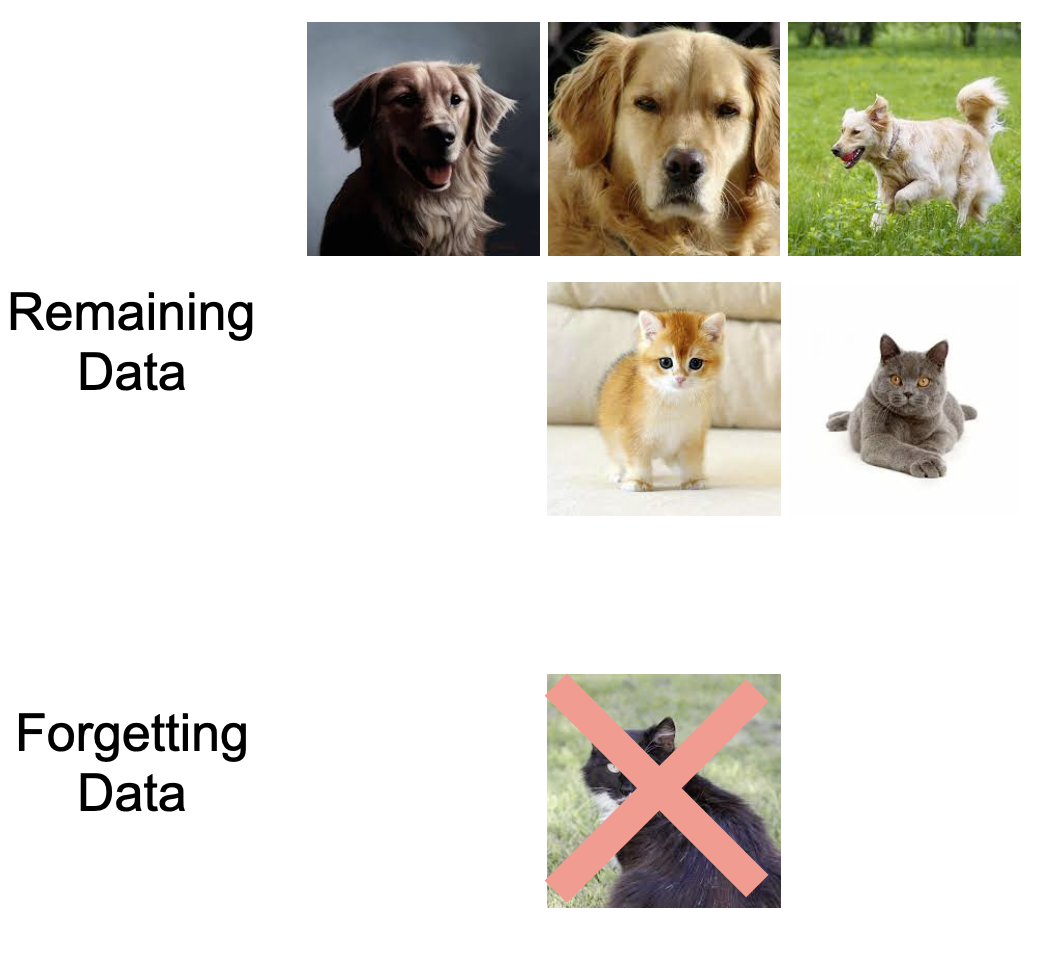}
        \end{minipage}}  
    \hspace{3mm}
    \subfigure[Conceptual independent forgetting and remaining data]{
        \begin{minipage}[htbp]{0.45\linewidth}
            \centering
            \includegraphics[width=\textwidth,height=\textwidth]{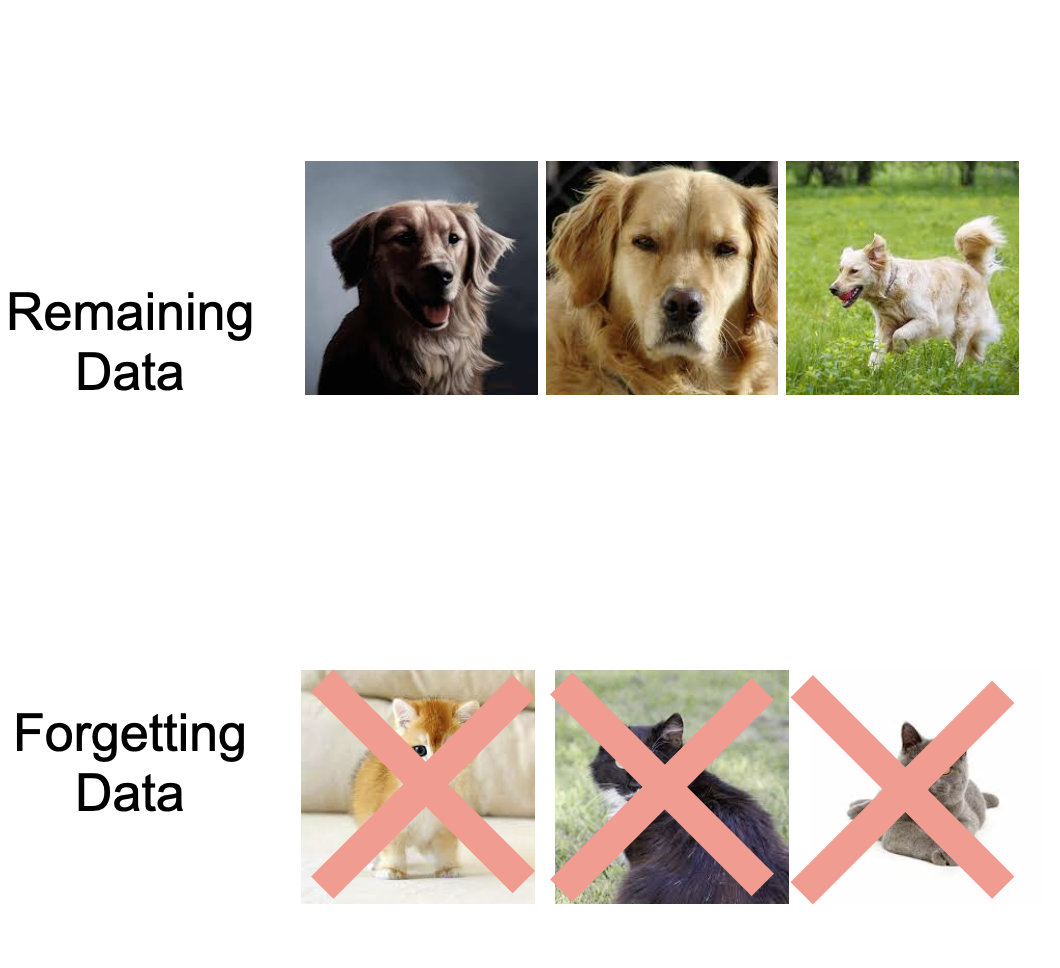}
        \end{minipage}%
    }
    \centering
    \vspace{-1em}
    \caption{Two different types of intertwining between forgetting and remaining data. For (a), the remaining data and forgetting data share the concept of ``cat''. For (b), the forgetting data and remaining data are from different concepts - ``cat'', and ``dog'' respectively.}
    \label{fig:intro}
    \vspace{-5mm}
\end{figure}

To illustrate, consider the example in Figure~\ref{fig:intro}(a), where the remaining data comprises images of ginger and grey cats, while the forgetting data includes an image of a black cat. In this example, the pre-unlearning model contains latent information about a black cat. When this information is completely removed, it may inadvertently affect the representation of cats in general, impacting the post-unlearning model's performance. Conversely, preserving an excessive amount of information about the cat may allow the post-unlearning model to excel at classifying black cats, resulting in insufficient unlearning. It is worth noting that preserving information about the forgetting data may also trigger the Streisand effect~\cite{Sunshine}, where the information in the forgetting data becomes identifiable in post-unlearning models. This raises significant concerns regarding privacy preservation~\cite{boundary, Sunshine}.

In this way, we highlight a crucial, newly identified challenge in machine unlearning: disentangling the intricately intertwining information between forgetting and remaining data, as failure to do so can lead to persistent issues of insufficient or excessive unlearning. The difficulty in accurate disentanglement arises from the varying degrees and types of intertwining that can exist, each representing distinct relationships between the forgetting data and the remaining data. These intertwinings can exhibit a wide spectrum of degrees and types, ranging from fully independent relationships, as illustrated in Fig.~\ref{fig:intro}(b), where disentanglement is unnecessary (an extreme example), to overlapping of concepts, such as the cat example in Fig.~\ref{fig:intro}(a), to more complex and nuanced interactions. Consequently, the question that arises is how to devise a universal disentangling method capable of effectively separating the intertwined forgetting and remaining data, while automatically adapting to the diverse relationships and degrees of intertwining.

To address this question, our paper introduces an innovative approach that integrates the principles of causal inference into the field of machine unlearning. By leveraging causal relationships, we endeavor to craft a versatile and adaptive disentangling method. Specifically, our approach begins by constructing causal graphs at three distinct levels: data, representation, and output, allowing us to account for a diverse range of information intertwining between forgetting data and remaining data. As a result, we reframe the unlearning problem as a dual task that encompasses both the removal of causal effects of forgetting data and the preservation of causal effects of remaining data through rigorous causal analysis. 

Building upon this analysis, we introduce a novel framework known as Causal Machine Unlearning (CaMU)\footnote{https://github.com/ShaofeiShen768/CaMU}. CaMU operates by concurrently aligning the representation and outputs of the forgetting data with those of its counterfactual data, effectively erasing its causal impact, while simultaneously maintaining the representation and outputs of remaining data in line with the pre-unlearning model to uphold its causal influence. To substantiate the effectiveness of CaMU in addressing the aforementioned challenges, we subject it to examination under two distinct scenarios: \emph{data point removal}, where some of the remaining data points belong to the same class as those in the forgetting data, indicating a substantial class overlap, and \emph{class removal}, where an entire class of data is eliminated, and the forgetting data and remaining data are distinctly separate, belonging to different classes.

The contributions of this paper can be summarised as follows:
\vspace{-2mm}
\begin{itemize}
\setlength\itemsep{-0.5mm}
    \item We pioneer the integration of causal inference into the realm of unlearning, thereby transforming the unlearning problem into a challenge of causal effect removal. This transition is accomplished through a meticulous analysis of the novel causal graph associated with the unlearning process in both pre-unlearning and post-unlearning models.
    \item We provide a comprehensive causal analysis of traditional unlearning, shedding light on the underlying causes of persistent issues such as residual latent information and performance degradation. 
    \item We propose the CaMU framework which works effectively for unlearning problems whether there is substantial overlap or independence between forgetting and remaining data. Through extensive empirical evaluations across diverse datasets and models, we demonstrate the superior performance of CaMU when compared to other methods.
    
\end{itemize}


\section{Preliminaries}

In this section, we formalize the deep model unlearning problem and propose causal graphs for the unlearning process on which we reframe the unlearning problem into a causal effect removal problem. The notations used throughout the paper are summarized in Table \ref{notiontable}.

\subsection{Deep Model Unlearning}\label{subsec:deep-model-unlearning}

We consider a supervised learning task on a dataset $D$ on which a training algorithm $A$ is applied to train the model $A(D)$. Given the forgetting data $F\subset D$, the unlearning algorithm $U$ is expected to remove the learned information on $F$ from $A(D)$. The model $A(D)$ after unlearning by $U$ is expected to perform similarly as the model which is retrained on the remaining data $R = D - F$, that is, $U(R, F, A(D))\approx A(R)$ when applied to instance $x$ sampled from the same distribution of $D$.

\subsection{Causal Graph For Unlearning}

\begin{figure}[t]
    \vspace{-2mm}
    \centering
    \subfigure[Three levels in causal graphs.]{
        \begin{minipage}[htbp]{0.3\linewidth}
            \centering        
            \includegraphics[width=\textwidth,height=\textwidth]{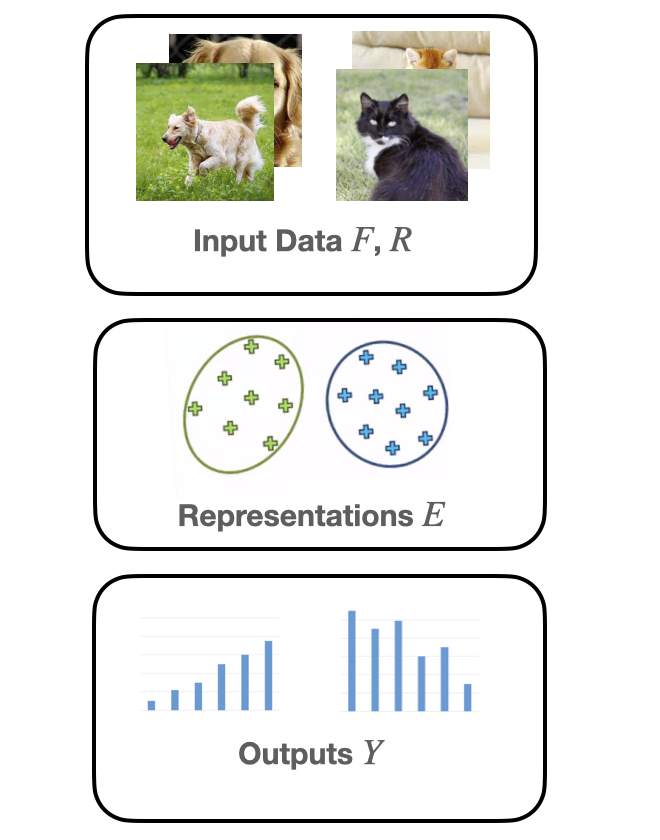}
        \end{minipage}%
    }%
    \hspace{2mm}
    \subfigure[Causal graph in pre-unlearning phase.]{
        \begin{minipage}[htbp]{0.25\linewidth}
            \centering
            \includegraphics[width=\textwidth,height=\textwidth]{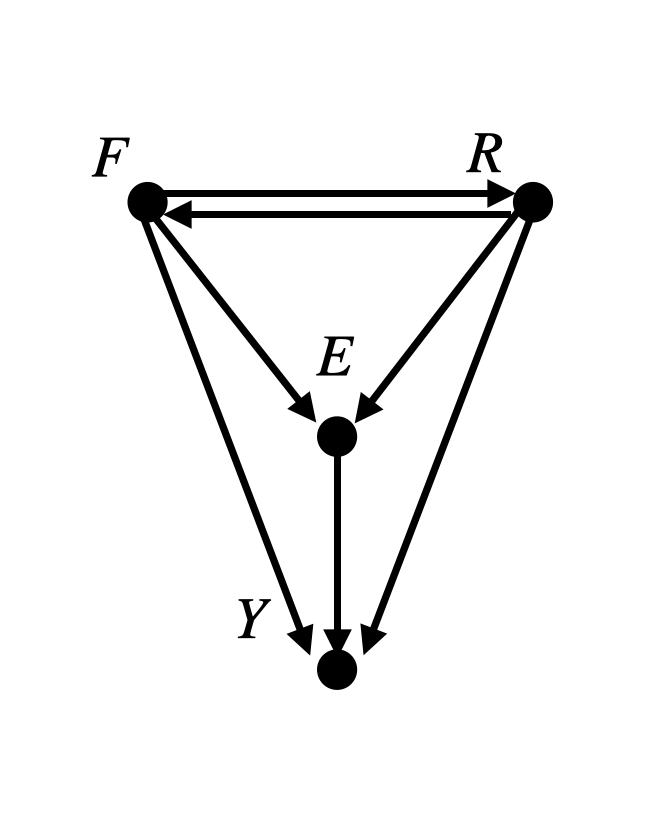}
        \end{minipage}%
    }
    \hspace{2mm}
    \subfigure[Causal graph in post-unlearning phase.]{
        \begin{minipage}[htbp]{0.25\linewidth}
            \centering
            \includegraphics[width=\textwidth,height=\textwidth]{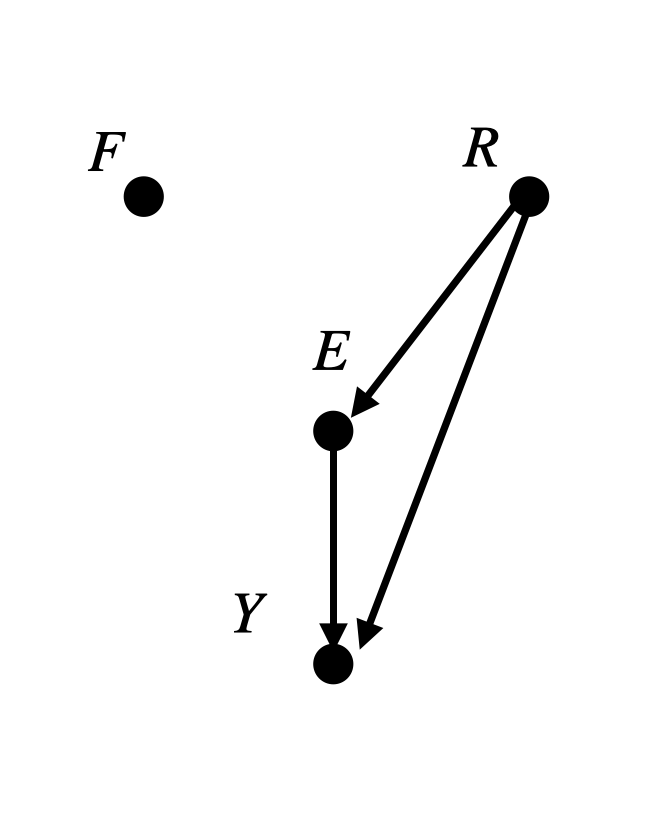}
        \end{minipage}%
    }%
    \centering
    \vspace{-1em}
    \caption{Causal graphs in pre-unlearning and post-unlearning phases. (a) presents three levels of causal variables and (b), and (c) denote the causal graphs of the pre-unlearning and post-unlearning phases respectively.}
    \label{fig:analysis}
    \vspace{-6mm}
\end{figure}

To understand the cause of residual information retention and performance degradation, we construct causal graphs \cite{causality} of the pre-unlearning model and the post-unlearning model. Inspired by \cite{DistillingCausalEffect}, we construct the causal graph using three levels of components: input data, representation, and output prediction, which are represented as variable nodes in the graphs shown in Fig. \ref{fig:analysis} in which the directed edges denote the causal relationships between the variable nodes.

Specifically, in Fig. \ref{fig:analysis}(b), $F \rightarrow E$ $\&$ $R \rightarrow E$ denote that in the pre-unlearning phase, the representation $E$ is extracted based on the forgetting data $F$ and remaining data $R$ through the feature extractors. Similarly, $F \rightarrow Y$ $\&$ $R \rightarrow Y$ denote the direct causal effect from $F$ and $R$ to the output distribution $Y$ while $E \rightarrow Y$ is the projection of representations $E$ to the outputs $Y$. The two distinct paths from $R$ and $F$ to $E$ and subsequently to $Y$ are present to illustrate their varying impacts on $Y$ due to the different distributions in $R$ and $F$. $F \rightarrow R$ $\&$ $R \rightarrow F$ denotes the intertwining information between $F$ and $R$, depicting the shared information between $R$ and $F$.

In the causal graph of the post-unlearning phase shown in Fig. \ref{fig:analysis}(c), $R \rightarrow E$ denotes the extraction of representation $E$ on the remaining data $R$ without any information about the forgetting data $F$.  $R \rightarrow Y$ $\&$ $R \rightarrow E  \rightarrow Y$ denotes the direct and indirect causal effect from the remaining data $R$ to the output distribution $Y$ of the post-unlearning model. In the unlearning described in Sec.~\ref{subsec:deep-model-unlearning}, the retraining of $A(R)$ does not use any forgetting data $F$. Therefore, the variable $F$ should not have any causal effect on other variables here.

\subsection{Reframing Unlearning In Causal View}

The unlearning objective requires transforming the causal graphs from Fig. \ref{fig:analysis}(b) to Fig. \ref{fig:analysis}(c). Specifically, given the training data $D$ consisting of the forgetting data $F$ and remaining data $R$, the model trained on $D$ can project the input data into representations $E$ and outputs $Y$. The causal relationships of the original model $A(D)$ can be denoted as Fig. \ref{fig:analysis}(b) while the causal relationships of the retrained model $A(R)$ are denoted in Fig. \ref{fig:analysis}(c). Then the unlearning problem can be viewed as the removal of causal paths $F\rightarrow E$ and $F \rightarrow Y$, which means the elimination of the direct influence of the forgetting data $F$ to the model $A(D)$, and the removal of causal paths $R\rightarrow F$ and $F\rightarrow R$, which means the disentanglement of information between forgetting and remaining data. In the meantime, the preservation of causal paths $R\rightarrow E$ and $R\rightarrow Y$ are required to retain the influence of remaining data.

\begin{table}[h]
\footnotesize
    \begin{center}
    \vspace{-3mm}
    \caption{Table of Notation}
    \vspace{1mm}
    \label{notiontable}
    \begin{tabular}{p{1.5cm} p{6cm} }
    \toprule
    Notation &  Explanation\\
    \midrule
    $A$ &  Learning algorithm\\
    $U$ &  Unlearning algorithm\\
    $P(\cdot)$& Probability\\
    $P(\cdot|\cdot)$& Conditional Probability\\\hline
    $D$ &  Training Data\\
    $F$ &  Forgetting data\\
    $R$ &  Remaining data\\
    $E$ & Extracted features in the pre-unlearning phase\\
    $Y$ & Output in the pre-unlearning phase\\\hline
    $s_f$& A forgetting example in $F$, $s_f = (x_f, y_f)$, whose random variables are $(X_F, Y_F)$\\
    $s_{r^*}$& A sampled remaining example in $R$, $s_{r^*} = (x_{r^*},y_{r^*})$, whose random variables are $(X_{R^*}, Y_{R^*})$\\
    $s_{f^*}$& A counterfactual example, $s_{f^*} = (x_{f^*},y_{f^*})$, whose random variables are $(X_{F^*}, Y_{F^*})$\\
    $s$& A tuple example, $s = (s_f, s_{r^*}, s_{f^*})$\\
    $\epsilon$ & Mask to generate the counterfactual example\\
    \hline
    $R^*$ &  Set of $s_{r^*}$\\
    $F^*$ &  Set of Counterfactual data\\
    $E^*$ &  Post-unlearning feature representation\\
    $Y^*$ &  Post-unlearning output space of model\\
    $S$ & Set of tuples $s$ \\\hline
    $g_o(\cdot)$ & Pre-unlearning model\\
    $g_u(\cdot)$& Post-unlearning model\\
    $\Tilde{E}$ & Pre-unlearning model's extracted feature representation\\
    $\Tilde{Y}$ & Pre-unlearning model's output\\
    $\ell(\cdot,\cdot)$ & Loss function\\
    \bottomrule
    \end{tabular}
    \end{center}  
    \vspace{-7mm}
\end{table}

\begin{figure}[h]
    \centering
    \subfigure[Original model]{
        \begin{minipage}[htbp]{0.29\linewidth}
            \centering        
            \includegraphics[width=\textwidth,height=\textwidth]{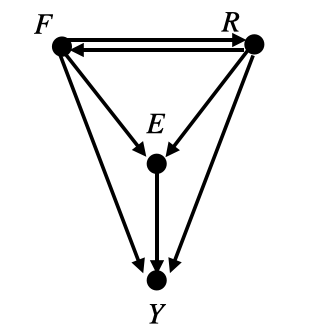}
        \end{minipage}%
    }%
    \hspace{2mm}
    \subfigure[Conventional unlearning]{
        \begin{minipage}[htbp]{0.29\linewidth}
            \centering
            \includegraphics[width=\textwidth,height=\textwidth]{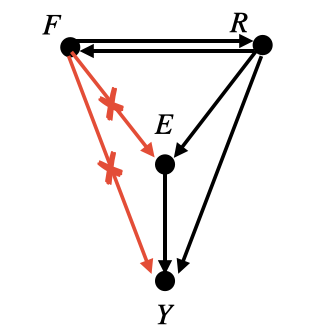}
        \end{minipage}%
    }
    \hspace{2mm}
    \subfigure[Causal unlearning]{
        \begin{minipage}[htbp]{0.29\linewidth}
            \centering
            \includegraphics[width=\textwidth,height=\textwidth]{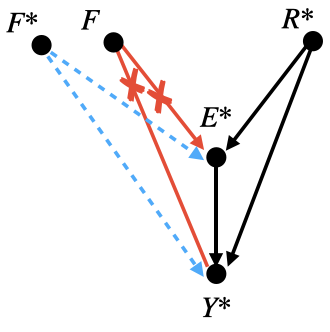}
        \end{minipage}%
    }%
    \centering
    \vspace{-2em}
    \caption{Comparison between conventional unlearning and causal unlearning}
    \label{fig:graph}
    \vspace{-5mm}
\end{figure}


\section{Causal Machine Unlearning}\label{s3}

To remove the causal effects of the forgetting data and preserve the causal effects of the remaining data, we propose Causal Machine Unlearning (CaMU) based on the causal analysis of the causal graph in Fig. \ref{fig:analysis}. In this section, we first provide a comparison between conventional unlearning methodology and causal-view unlearning. Subsequently, we introduce details of CaMU.

\subsection{Conventional Unlearning}

In previous works on deep model unlearning \cite{unroll,Amnesiac,scrub}, the objective focuses on removing the impacts of the forgetting data $F$ on the model predictions $Y$, which can be represented as the paths $F\rightarrow Y$ and $F \rightarrow E \rightarrow Y$ in Fig. \ref{fig:graph} (a). The general objective can be regarded as removal or minimization the causal effect $F \rightarrow Y$ and $F \rightarrow E \rightarrow Y$ as shown in Fig. \ref{fig:graph} (b).

Without disentangling the intertwining information between $F$ and $R$, the removal of the above two paths will also affect the path $F \rightarrow R \rightarrow E$ and $F \rightarrow R \rightarrow Y$. As a consequence, a thorough removal of the two paths will damage the information from the remaining data $R$. Conversely, if the unlearning algorithm is designed to maintain the integrity of information on the remaining data $R$, the latent information from the forgetting data $F$ may be inadvertently preserved through the causal path $R \rightarrow F$.

\subsection{Counterfactual Data Preparation} To disentangle the intertwining information between $F$ and $R$, we first prepare some counterfactual data based on them. Counterfactual data act as a mechanism to recalibrate the model's comprehension, simulating the scenario as if the data designated for forgetting had never been incorporated initially. By substituting the forgetting data with counterfactuals that closely resemble the remaining data, the model is prompted to discard the learned representations associated with the forgetting data. Concurrently, the employment of randomly drawn labels for these counterfactuals aims to disrupt and erase the specific output patterns learned from the forgetting data. This dual approach ensures a thorough and effective unlearning process, targeting both the representation and output dimensions.

Specifically, for each sample $s_f$ within the forgetting data $F$, we commence by uniformly sampling $s_{r^*}=(x_{r^*},y_{r^*})$ from the remaining dataset, where $x_{r^*}$ represents the raw features and $y_{r^*}$ the label. Following this, we generate a random mask $\epsilon$ of identical dimensions to $x_{r^*}$ and comparable scale to its entries\footnote{Given that the entries in $x_{r^*}$ have undergone normalization, each element within the mask is uniformly sampled from the range $[0,1]$.}. Concurrently, we select a random label from the set of labels, excluding the label of the current forgetting sample. By adding together the random mask $\epsilon$ with the remaining sample $x_{r^*}$, and annotating it with the newly sampled label, we construct a counterfactual sample. Each tuple comprising the forgetting sample, the remaining sample, and the counterfactual sample is then incorporated into the joint dataset $S$. This process is replicated for every $s_f \in F$. The data preparation procedure is summarized in Algorithm~\ref{alg:Framework1}.


\begin{algorithm}[htb] 
\caption{Data Preparation} 
\label{alg:Framework1} 
\begin{algorithmic}[1]
\REQUIRE ~~\\ 
The forgetting data, $F$;\\
The remaining data, $R$;
\ENSURE ~~\\
The joined dataset $S = \{(s_f, s_{f^*}, s_{r^*})|\forall s_f\in F\}$;
\STATE {\bfseries for} each forgetting sample $s_f = (x_f, y_f) \in F$:\\
\STATE \quad Sample $s_{r^{*}} = (x_{r^*}, y_{r^*})\in R$ uniformly at random;
\STATE \quad Sample a mask $\epsilon$ uniformly at random from $[0,1]$, ensuring $\mbox{size}(\epsilon) = \mbox{size}(x_{r^*})$; 
\STATE \quad Sample $y^*\neq y_{f}$ uniformly at random;
\STATE \quad $s_{f^{*}} = ({x_{r^{*}} + \epsilon, y^*})$;
\RETURN $S = \{(s_f, s_{f^*}, s_{r^*})|\forall s_f\in F\}$;
\end{algorithmic}
\end{algorithm}
\vspace{-3mm}

\subsection{Causal Unlearning}
To solve the issues that are caused by the intertwining information between $F$ and $R$, we propose CaMU to reduce the latent information from forgetting data and performance degradation on remaining data. The desired causal graph is presented in Fig. \ref{fig:graph} (c), in which $F^*$ contains all the counterfactual data $s_{f^*}$ generated from the tuples in $S$, and $E^*$ and $Y^*$ are the desired post-unlearning extracted features and output respectively. Similarly, $R^*$ contains all 
remaining data $s_{r^*}$ from the tuples in $S$. In Fig. \ref{fig:graph} (c), the overall objective is to disentangle $F$ and $R^*$, remove the causal paths $F\rightarrow E^*$ and $F\rightarrow Y^*$, but preserve valuable information through the counterfactual samples $F^*$ by constructing causal paths $F^*\rightarrow E^*$ and $F^*\rightarrow Y^*$.  

\noindent\textbf{Remove $\boldsymbol{{F}\rightarrow Y^*}$.} To disconnect the causal pathway $F\rightarrow Y^*$ while concurrently dissolving the interdependence between $F$ and $R$, we employ counterfactual data randomly generated from $F$ and $R^*$. Once $F^*$ is established, $F$ and $R^*$ become independent. Hence, at this juncture, our primary focus is the interaction between $F^*$ and $R^*$ and their respective influences on $E^*$ and $Y^*$.

Fundamentally, our aim is that $F$ will mirror the output of the randomly generated $F^*$ for effective unlearning, which is mathematically articulated as
\begin{align} \label{eq3}
    \min\quad d(P(Y^{*}|F), P(Y^{*}|F^{*})),
\end{align}
where $d(\cdot,\cdot)$ denotes the distance or divergence measure between two probability distributions. It is important to note that
\begin{align} \label{eq3-1st-item}
    P(Y^{*}|F)  = \sum_{E^{*}} P(Y^{*}|F,E^{*})P(E^{*}|F),
\end{align}and
\begin{align} \label{eq3-2nd-item}
    P(Y^{*}|F^{*})=&\sum_{E^{*}} P(Y^{*}|F^*,E^{*})P(E^{*}|F^*) \\ \notag
    = &\sum_{E^{*}} P({Y}_{F^{*}})P(E^{*}|F^{*}),
\end{align}
where we assume  $P(Y^{*}|F^*,E^{*})$ can be perfectly approximated by  $P({Y}_{F^{*}})$, the groundtruth of data in $F^*$. By leveraging Eqs.~\ref{eq3-1st-item} and~\ref{eq3-2nd-item}, we can indirectly achieve Eq.~\ref{eq3}, which is typically challenging to estimate, through
\begin{align}
    \min \quad d(P(E^{*}|F), P(E^{*}|F^{*})),\label{eq5}
\end{align}
and
\begin{align}
    \min \quad d(P(Y^{*}|F,E^{*}),P({Y}_{F^{*}})).\label{eq6}
\end{align}

\noindent\textbf{Preserve $\boldsymbol{R\rightarrow Y^*}$.} 
Another aspect of unlearning is to preserve the effect on the remaining data, which is achieved by
\begin{align}\label{eq7} 
    \min \quad d(P(Y^*|R^{*}), P(\Tilde{Y}|R^{*})),
\end{align}
where $\Tilde{Y}$ is the output label of the pre-unlearning model for data in $R^*$. Similarly,
\begin{align}\label{eq7.1} 
   P(Y^*|R^{*}) =\sum_{E^{*}} P(Y^{*}|R^{*},E^{*})P(E^{*}|R^{*}),  
\end{align}
and
\begin{align}\label{eq7.2} 
    P(\Tilde{Y}|R^{*}) 
    &= \sum_{E^{*}} P(\Tilde{Y}|R^{*},E^{*})P(E^{*}|R^{*})\\ \notag
    &= \sum_{\Tilde{E}} P({Y}_{R^{*}})P(\Tilde{E}|R^{*}),
\end{align}
in which $\Tilde{E}$ is the representation learned by the pre-unlearning model, which should be preserved in $E^*$ on $R^*$, and $Y_{R^{*}}$ represents the groundtruth labels, which is the objective for the pre-unlearning model to map. Similarly, by leveraging Eqs.~\ref{eq7.1} and~\ref{eq7.2}, we can indirectly achieve Eq.~\ref{eq7} through 
\begin{equation}
    \min \quad d(P(E^{*}|R^{*}), P(\Tilde{E}|R^{*}) ) \label{eq9},
\end{equation}
and
\begin{equation}
    \min \quad d(P(Y^{*}|R^{*},E^{*}), P({Y}_{R^{*}}|R^{*},\Tilde{E})). \label{eq10}
\end{equation}

\begin{algorithm}[t] 
\caption{Causal Machine Unlearning Algorithm: CaMU} 
\label{alg:Framework2} 
\begin{algorithmic}[1]
\REQUIRE ~~\\ 
The joint dataset, $S$;\\
Epochs for causal effect removal $T$\\
The pre-unlearning deep model $g_o$\\
\ENSURE ~~\\
The post-unlearning model $g_u$;\\
\STATE Initialize $g_u = g_o$
\STATE {\bfseries for} $t$ in $range(T)$:\\
\STATE \quad Update $g_u$ by Eq.~\ref{eq15-s}
\STATE \quad Update $g_u$ by Eq.~\ref{eq16-s}
\RETURN $g_u$;
\end{algorithmic}
\end{algorithm}
\vspace{-2mm}

\noindent\textbf{Algorithm}
The overall objective of our algorithm is to optimize Eqs.~\ref{eq5},~\ref{eq6},~\ref{eq9}, and~\ref{eq10}. Among them, Eqs.~\ref{eq5} and~\ref{eq9} minimize the discrepancy between the distributions on the extracted features, so we use the KL-divergence as $\mathrm{KL}(\cdot,\cdot)$ and minimize the following objective in each batch of data
\begin{align}\label{eq15}
    \ell_{\mathrm{KL}} = &\mathrm{KL}(P(E^{*}|F),P(E^{*}|F^{*})) +\\ \notag
    &\mathrm{KL}(P(E^{*}|R^{*}),P(\Tilde{E}|R^{*})).\notag
\end{align}

Eqs.~\ref{eq6} and~\ref{eq10} estimate the differences of the probabilities in the output distribution, so we use the cross-entropy ($\mathrm{CE}$), and organize these two equations into one objective
\begin{align}\label{eq16}
    \ell_\mathrm{CE} = &\mathrm{CE}(P(Y^{*}|F,E^{*}),P({Y}_{F^{*}})) +\\\notag
    &\mathrm{CE}(P(Y^{*}|R^{*},E^{*}),P({Y}_{R^{*}})).  \notag
\end{align}

Denoted by $g_o(\cdot)$ the pre-unlearning classification model, which gives a representation of $g_o^e(\cdot)$, and $g_u(\cdot)$ the post-unlearning classification model, which gives the corresponding representation by $g_u^e(\cdot)$. For a tuple $s = (s_f, s_{r^*}, s_{f^*})\in S$ where $s_f = (x_f, y_f)$, $s_{r_*} = (x_{r^*}, y_{r^*})$ and $s_{f^*} = (x_{f^*},y_{f^*})$, we have the learning objective of
\begin{align}\label{eq15-s}
    \min_{g_u}\ell_{\mathrm{KL}} (S)&= \mathrm{KL}(g_u^e(X_F), g^e_u(X_{F^*})) + \\ \notag &\quad\quad\quad\mathrm{KL}(g_u^e(X_{R^*}), g_o^e(X_{R^*}))
\end{align}
and
\begin{align}\label{eq16-s}
    \min_{g_u} \ell_\mathrm{CE} (S)= &\frac{1}{|F|}\sum_F \mathrm{CE}(g_u(x_f), y_{f^*}) +\\\notag
    &\quad\quad\frac{1}{|R^*|}\sum_{R^*}\mathrm{CE}(g_u(x_{r^*}), y_{r^*}).  \notag
\end{align}

The pseudo-code of CaMU is provided in Algorithm \ref{alg:Framework2}. In CaMU, we use the joint dataset $S$ and the pre-unlearning model $g_o$ as input. We set a cloned $g_o$ to initialize the post-unlearning model $g_u$. Then $g_u$ is optimized on $S$ via minimizing $l_\mathrm{KL}$ and $l_\mathrm{CE}$ alternatively.

\section{Experiments}

In this section, we conduct experiments to answer three research questions to evaluate CaMU:
\vspace{-2mm}
\begin{itemize}
    \setlength\itemsep{-1mm}
    \item \textbf{RQ1}: How does the proposed CaMU perform on different unlearning tasks when the distribution of the forgetting data and remaining data are overlapping or independent, as compared with the state-of-the-art unlearning methods? 
    \item \textbf{RQ2}: How do the operations on each part of the proposed causal graph affect the effectiveness of the proposed CaMU?
    \item \textbf{RQ3}: How stable is the post-unlearning model via CaMU framework in terms of the forgetting data and the remaining data performances while relearning the remaining data?
\end{itemize}
\vspace{-1mm}
\subsection{Settings}

\subsubsection{Datasets and Models}
To validate the effectiveness of the CaMU framework, we conduct experiments on four different datasets: \textbf{Digit} \cite{mnist}, \textbf{Fashion} \cite{fashion}, \textbf{C10} \cite{cifar} and \textbf{C100} \cite{cifar}. Among them, Digit and Fashion are both MNIST datasets. Detailedly, Digit consists of images of handwritten digits from 0 to 9 and Fashion contains ten common clothes. Both MNIST datasets contain 60,000 training samples and 10,000 test samples. C10 (CIFAR-10) and C100 (CIFAR-100) contain images of ten and one hundred, respectively, common animals or items. Both CIFAR datasets contain 50,000 training samples and 10,000 test samples. For the two MNIST datasets, we use a convolutional neural network (\textbf{CNN}) with two convolutional layers \cite{cnn} and three linear layers while for the two CIFAR datasets, we choose an \textbf{18-layer ResNet} backbone\cite{resnet} without pre-trained parameters.



\subsubsection{Baselines}

We compare the performance of \textbf{CaMU} with seven baseline results including \textbf{Retrain} which are the golden standards of unlearning, and six state-of-the-art unlearning works on deep models: \textbf{NGrad}, \textbf{Boundary}~\cite{boundary}, \textbf{T-S}~\cite{ts}, \textbf{SCRUB}~\cite{scrub}, \textbf{SISA}~\cite{sisa}, and \textbf{Unroll}~\cite{unroll}. Detailed descriptions of these methods can be found in the appendix.

\subsubsection{Evaluation Setting}
Firstly, to prove the efficiency and applicability of the proposed CaMU on overlapped and independent forgetting data and remaining data settings, we set up two situations of experiments: (1) random data removal where we randomly select a group training data to remove and the distribution of the forgetting data and remaining data should have many overlaps; (2) class removal which consists of label unlearning where we regard the several classes of data as the forgetting data and regard the left classes as the remaining data. The distribution of different classes of data should have few overlaps and be independent of each other. As for the evaluations, we compare the performances of CaMU with the baselines in terms of the following seven metrics:

\noindent \quad Used for \textbf{random data removal} experiments:
\vspace{-2.5mm}
\begin{itemize}
\setlength\itemsep{-1mm}
    \item \boldmath$R_{tr}$ (accuracy of the remaining data): Closer values to the retrained model indicate better performance.
    \item \boldmath$F_{tr}$ (accuracy of the forgetting data): Closer values to the retrained model indicate better performance.
\end{itemize}
\vspace{-1.5mm}
\noindent \quad Used for \textbf{class removal} experiments:
\vspace{-2.5mm}
\begin{itemize}
\setlength\itemsep{-1mm}
    \item \boldmath$R_{ts}$ (accuracy of test data on remaining class): Higher performances mean better performances.
    \item \boldmath$F_{ts}$ (accuracy of test data on forgetting class): Lower performances mean better performances.
\end{itemize}
\vspace{-1.5mm}
\noindent \quad Used for \textbf{both} experiments:
\vspace{-2.5mm}
\begin{itemize}
\setlength\itemsep{-1mm}
    \item \boldmath$Ts$ (accuracy of test data): Higher performances mean better performances.
    \item \boldmath$M$ (membership inference attack): Closer values to the retrained model indicate better performance.
    \item \textbf{Unlearning time}: Less time stands for higher algorithm efficiency.
\end{itemize}
\vspace{-1.5mm}



All the experiment results are the average of five rounds of experiments using the same random seeds from 0 to 4. Further details of experiment settings and hyperparameters are described in the appendix. 

\subsection{Performance Comparison (RQ1)}
\subsubsection{Effectiveness Comparison}

The experiments start with the unlearning effectiveness comparison of random data removal and class removal, where we aim to remove 10\% random data and class 0 respectively. We compare the training and test accuracy of CaMU and other baselines to evaluate whether the unlearning algorithm can achieve the goals of unlearning and avoid latent information and performance degradation compared with the retrain-from-scratch models. The comparison results of random data removal are summarized in Table \ref{tab:random1}, \ref{tab:class1} and \ref{tab:mia}, where we have the following insights.\\\textbullet\quad In the experiments of random data removal, the proposed CaMU can reach the second-highest test accuracies ${Ts}$ on the two MNIST datasets apart from the NGrad and Unroll which show few differences between $R_{tr}$ and $F_{tr}$ and indicates failures on the data removal. In addition, the forgetting data accuracies $F_{tr}$ of CaMU are also the closest ones to the $F_{tr}$ of retraining models. On the two CIFAR datasets, the removal becomes more difficult because of the more complex images and model structures. CaMU can retain the highest remaining data accuracies and test accuracies. In addition, CaMU can also reach the closest $F_{tr}$ to the retrained models, except SISA on CIFAR100, which fails on forgetting. \\\textbullet\quad In the experiments of class removal, CaMU can always reach zero $F_{ts}$ on all four datasets, which indicates that CaMU has achieved class removal tasks in terms of prediction performances. Furthermore, CaMU can always reach the highest $R_{ts}$ on all four datasets which demonstrates the performance degradation has been minimized. \\\textbullet\quad In terms of the successful rate of MIA, the CaMU shows competitive results with other baselines. Therefore, to comprehensively compare the unlearning algorithms, we calculate the average absolute differences between the post-unlearning models and the retrained models by comparing the effectiveness metrics. The post-unlearning models of CaMU can achieve a 4.09 percentage average difference between the two removal tasks, which ranked second on all methods.

\subsubsection{Efficiency Comparison}

The time cost of CaMU mainly accounts for the optimization of Eq.~\ref{eq15} and Eq.~\ref{eq16}, where we construct the same number of counterfactual samples and remaining samples as the number of forgetting samples. Therefore, the total time cost of the removal task mainly depends on the size of the forgetting data and the type of task. Table \ref{tab:time} presents the time cost of CaMU and other baselines. In the experiments of random data removal and class removal on the two MNIST datasets, CaMU requires the second lowest time costs to finish the removal task, which is only higher than the time costs of Unroll. This is because Unroll recovers all the gradients of the forgetting data without any optimization approaches. In the experiments on CIFAR datasets, CaMU can achieve the highest time efficiencies even compared with Unroll in the class removal task on CIFAR10 while CaMU can reach the third highest in the data removal tasks on CIFAR10. Finally, in the class removal on CIFAR100, we repeatedly select the forgetting data to construct the joint dataset considering that the size of forgetting data is only 500. Thus, the time cost is a bit higher than other methods.

\subsection{Causal Effect Analysis (RQ2)}

We next conduct some ablation studies to understand how the operations (remove or preserve causal path) on the proposed causal graph affect the effectiveness of the proposed CaMU. Compared with the conventional unlearning methods, CaMU adds two parts of causal paths in Fig. \ref{fig:graph} (c): (1) $F^{*}\rightarrow E^{*}$ and $F^{*}\rightarrow Y^{*}$ and (2) $R^{*}\rightarrow E^{*}$. The first part denotes the counterfactual samples to remove the causal effect from the forgetting data $F$ to the post-unlearning model $E^{*}$ and $Y^{*}$ and the second one stands for the alignment of representations of the remaining samples $R^{*}$. Consequently, we conduct ablation studies on three other settings: \\\textbullet\quad Finetuning models only with the remaining data $R^{*}$ with the two added parts in Fig. \ref{fig:graph} (c). \\\textbullet\quad Finetune the model by both $R^{*}\rightarrow E^{*}$ and $R^{*}\rightarrow Y^{*}$ without the counterfactual data $F^*$. \\\textbullet\quad Remove causal effect from the forgetting data $F^{*}$ and preserve the model performances by finetuning without $R^{*}\rightarrow E^{*}$.

\begin{table}[h]
\tiny
\centering
\vspace{-5mm}
\caption{Performances of 10\% data removal task (\%). $\star$ means the best results and $\dagger$ stands for the second best. The two notations have the same meanings in the following tables.}
\label{tab:random1}
\vspace{-4mm}
\setlength\tabcolsep{3pt}
\renewcommand{\arraystretch}{1.25} 
\begin{center}
\begin{tabular}{p{0.5cm}<{\centering} |p{0.5cm}<{\centering} p{0.7cm}<{\centering}p{0.7cm}<{\centering}p{0.7cm}<{\centering}p{0.7cm}<{\centering}p{0.7cm}<{\centering}p{0.7cm}<{\centering}p{0.7cm}<{\centering}p{0.7cm}<{\centering}}  \toprule
Data & Metric & Retrain& NGrad& {Boundary}& {T-S}& {SCRUB} & {SISA}& {Unroll}& {CaMU}\\  \midrule

 \multirow{3}{*}{\rotatebox{90}{Digit}} & $R_{tr}$& {99.59}& $\text{99.48}^{\star}$& {97.60}& $\text{99.37}^{\dagger}$& {99.25}& {89.38}& {97.33} & {99.24} \\ 
                        & $F_{tr}$& {98.79}& {99.37}& {97.63}& {95.68}& $\text{99.22}^{\dagger}$& {89.44}& {97.30} & $\text{99.12}^{\star}$ \\
                        & $Ts\uparrow$& {98.93}& $\text{98.93}^{\star}$& {97.07}& {98.16}& {99.00}& {89.41}& {96.85} & $\text{98.98}^{\dagger}$ \\\midrule
 \multirow{3}{*}{\rotatebox{90}{Fashion}} & $R_{tr}$& {94.41}& $\text{93.61}^{\dagger}$& {50.12}& {92.96}& {90.91}& {84.53}& $\text{93.64}^{\star}$ & {92.51} \\ 
                        & $F_{tr}$& {91.00}& {92.79}& {49.86}& {88.26}& $\text{90.83}^{\dagger}$& {84.61}& {93.34} & $\text{90.96}^{\star}$ \\
                        & $Ts\uparrow$& {90.51}&  {89.60}& {48.93}& {89.30}& {89.21}&{82.93}& $\text{89.62}^{\star}$ & $\text{88.69}^{\dagger}$ \\\midrule
 \multirow{3}{*}{\rotatebox{90}{C10}} & $R_{tr}$& {84.25}& {75.38}& {55.80}& $\text{78.61}^{\dagger}$& {27.41}& {73.30}& {77.65} & $\text{82.19}^{\star}$ \\ 
                        & $F_{tr}$& {79.91}& {73.44}& {55.33}& {75.71}& {26.81}& {73.38}& $\text{77.07}^{\dagger}$ & $\text{81.64}^{\star}$ \\
                        & $Ts\uparrow$& {87.49}& {82.33}& {64.87}& $\text{85.42}^{\dagger}$& {30.37}& {66.27}& {83.73} & $\text{87.32}^{\star}$ \\\midrule
 \multirow{3}{*}{\rotatebox{90}{C100}} & $R_{tr}$& {66.78}& {52.47}& {36.70}& {54.82}& {8.36}& $\text{55.22}^{\dagger}$& {50.40} & $\text{62.77}^{\star}$ \\ 
                        & $F_{tr}$& {55.40}& {47.67}& {36.71}& {49.55}& {8.39}& $\text{55.66}^{\star}$& {49.26} & $\text{60.92}^{\dagger}$ \\
                        & $Ts\uparrow$& {63.54}& {55.09}& {42.16}& $\text{58.36}^{\dagger}$& {8.79}& {37.92}& {54.11} & $\text{62.79}^{\star}$ \\
\bottomrule
\end{tabular}
\vspace{-8mm}
\end{center}

\end{table}

\begin{table}[h]
\tiny
\centering
\vspace{-5mm}
\caption{Performances of class 0 removal task (\%)}
\label{tab:class1}
\vspace{-4mm}
\setlength\tabcolsep{3pt}
\renewcommand{\arraystretch}{1.25} 
\begin{center}
\begin{tabular}{p{0.5cm}<{\centering} |p{0.5cm}<{\centering} p{0.7cm}<{\centering}p{0.7cm}<{\centering}p{0.7cm}<{\centering}p{0.7cm}<{\centering}p{0.7cm}<{\centering}p{0.7cm}<{\centering}p{0.7cm}<{\centering}p{0.7cm}<{\centering}}  \toprule
Data & Metric & Retrain& NGrad& {Boundary}& {T-S}& {SCRUB} & {SISA}& {Unroll}& {CaMU}\\ \midrule

 \multirow{2}{*}{\rotatebox{90}{Digit}} & $R_{ts}\uparrow$& {98.81}& {98.86}& {98.59}& {61.31}& {99.02}& $\text{99.10}^{\dagger}$& {97.05} & $\text{99.22}^{\star}$ \\ 
                        & $F_{ts}\downarrow$& {0}& {79.76}& {95.63}& {0.16}& {95.41}& $\text{0}^{\star}$& {79.63} & $\text{0}^{\star}$ \\\midrule

  \multirow{2}{*}{\rotatebox{90}{Fash-}\rotatebox{90}{ion}} & $R_{ts}\uparrow$ & {92.66}& {89.70}& {86.04}& {91.84}& {91.40}& $\text{92.14}^{\dagger}$& {88.72} & $\text{92.66}^{\star}$\\ 
                        & $F_{ts}\downarrow$& {0}& {0.92}& {1.68}& {21.16}& {0.42}& $\text{0}^{\star}$& {0.4} & $\text{0}^{\star}$ \\\midrule
 \multirow{2}{*}{\rotatebox{90}{C10}} & $R_{ts}\uparrow$& {87.01}& {57.55}& {83.33}& $\text{86.47}^{\dagger}$& {32.93}& {73.52}& {84.26} & $\text{87.16}^{\star}$ \\ 
                        & $F_{ts}\downarrow$& {0}& $\text{0}^{\star}$& {1.0}& {6.12}& $\text{0}^{*}$& $\text{0}^{*}$& $\text{0}^{\star}$ & $\text{0}^{\star}$ \\\midrule
 \multirow{2}{*}{\rotatebox{90}{C100}} & $R_{ts}\uparrow$& {61.08}& {52.51}& {41.33}& $\text{59.46}^{\star}$& {7.96}& {38.27}& {31.69} & $\text{56.52}^{\dagger}$ \\ 
                        & $F_{ts}\downarrow$& {0}& {6.0}& {1.8}& {22.40}& $\text{0}^{\star}$ & $\text{0}^{\star}$& $\text{0}^{\star}$ & $\text{0}^{\star}$ \\
\bottomrule
\end{tabular}
\vspace{-8mm}
\end{center}
\end{table}

\begin{table}[h]
\tiny
\centering
\vspace{-5mm}
\caption{Attack success rate comparisons in MIA}
\label{tab:mia}
\vspace{-4mm}
\setlength\tabcolsep{3pt}
\renewcommand{\arraystretch}{1.25} 
\begin{center}
\begin{tabular}{p{0.5cm}<{\centering} | p{0.9cm}<{\centering} | p{0.7cm}<{\centering}p{0.6cm}<{\centering}p{0.7cm}<{\centering}p{0.6cm}<{\centering}p{0.6cm}<{\centering}p{0.6cm}<{\centering}p{0.6cm}<{\centering}p{0.6cm}<{\centering}}  \toprule
Type &  Data &  Retrain& NGrad& {Boundary}& {T-S}& {SCRUB} & {SISA}& {Unroll}& {CaMU}\\  \midrule
 \multirow{4}{*}{\rotatebox{90}{Data}} & {Digit}& {49.58}& {49.61}& {49.59}& {41.11}& {49.00}& {50.00}& {49.47} & {47.91} \\
 ~& {Fashion} & {49.97}& {50.19}& {49.97}& {45.00}& {50.22}& {50.00}& {49.63} & {48.14} \\
~&{C10}& {57.96}& {57.16}& {60.30}& {56.30}& {55.28}& {50.12}& {57.60} & {54.10} \\
 ~&{C100}& {59.31}& {58.16}& {60.00}& {56.92}& {55.53}& {50.12}& {58.71} & {53.33} \\\midrule
  \multirow{4}{*}{\rotatebox{90}{Class}} & {Digit}& {26.49}& {39.71}& {38.51}& {35.83}& {32.04}& {50.12}& {40.16} & {23.62} \\
 ~& {Fashion} &  {38.24}& {37.64 }& {39.06}& {24.82}& {33.84}& {50.00}& {40.61} & {35.24} \\ 
~&{C10}& {67.76}& {50.46}& {61.22}& {44.95}& {50.59}& {50.12}& {67.59} & {63.58} \\
 ~&{C100}& {61.96}& {63.74}& {74.98}& {70.92}& {53.99}& {50.59}& {62.91} & {72.36} \\\midrule
 \multicolumn{2}{c}{Average Diff $\downarrow$}& {-}&  {4.38} & {4.43}& {8.89}& {5.30}& {10.23}& $\text{2.32}^{\star}$& $\text{4.22}^{\dagger}$ \\ 
\bottomrule
\end{tabular}
\vspace{-6mm}
\end{center}
\end{table}

\begin{figure}[t]
    \centering
    \vspace{-4mm}
    \subfigure[Digit]{
        \begin{minipage}[htbp]{0.44\linewidth}
            \centering
            \includegraphics[width=\textwidth,height=0.9\textwidth]{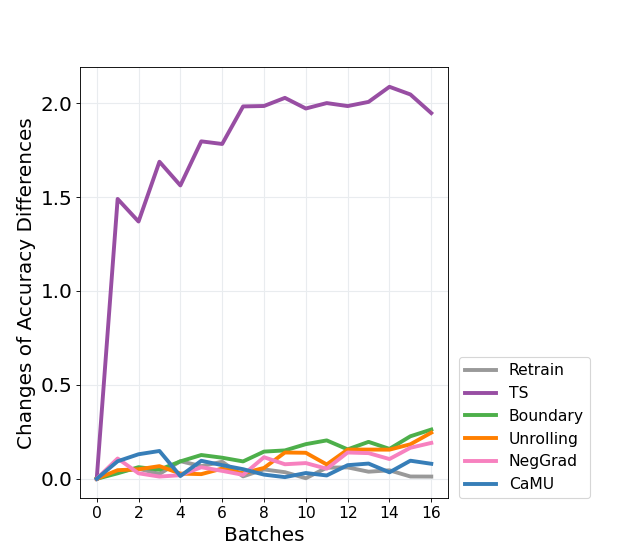}
        \end{minipage}%
    }
    \subfigure[Fashion]{
        \begin{minipage}[htbp]{0.44\linewidth}
            \centering
            \includegraphics[width=\textwidth,height=0.9\textwidth]{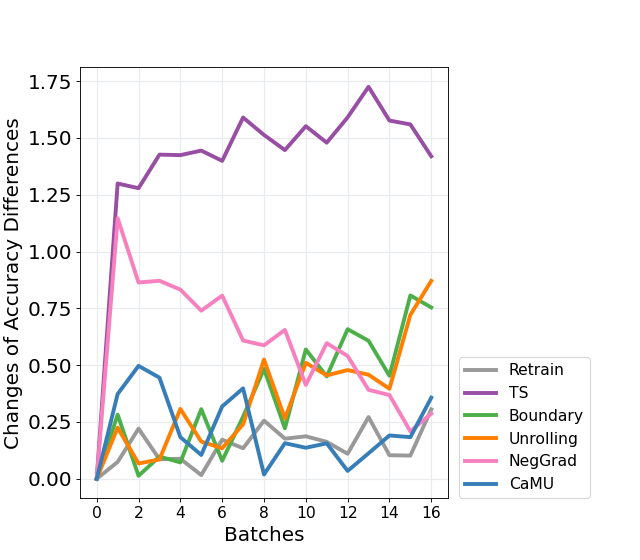}
        \end{minipage}%
    }%
    \vspace{-4mm}
    \subfigure[CIFAR10]{
        \begin{minipage}[htbp]{0.44\linewidth}
            \centering
            \vspace*{-0.4cm}
            \includegraphics[width=\textwidth,height=0.9\textwidth]{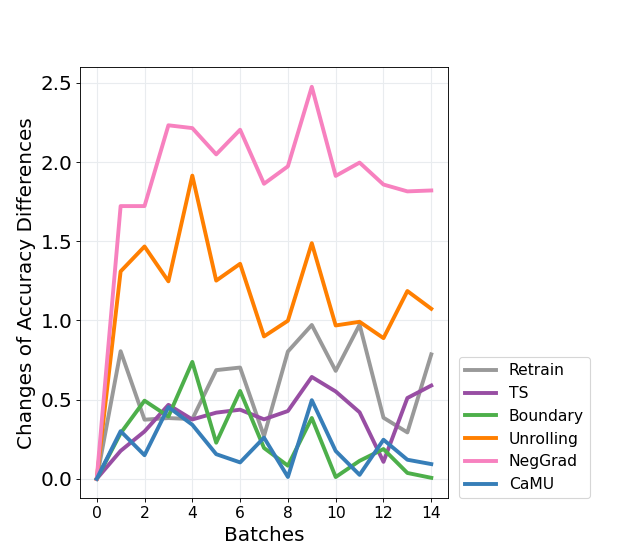}
        \end{minipage}%
    }%
    \subfigure[CIFAR100]{
        \begin{minipage}[htbp]{0.44\linewidth}
            \centering
            \vspace*{-0.4cm}
            \includegraphics[width=\textwidth,height=0.9\textwidth]{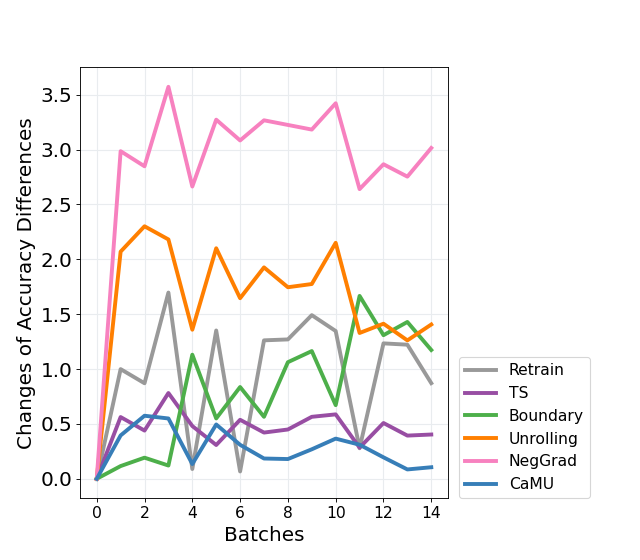}
        \end{minipage}%
    }%
    \centering
    \vspace{-1em}
    \caption{Changes of $R_{tr} - F_{tr}$ during relearning on the random data removal task}
    \label{fig:comparedata}
    \vspace{-6mm}
\end{figure}

Table \ref{tab:abl1} and Table \ref{tab:abl2} demonstrate the results of ablation studies in the above three settings and the entire CaMU algorithm. In the first setting, Finetune can always reach the closest MIA to the results of the retrained models because Finetune does not impair the model. Furthermore, compared with finetuning, the second setting where we add causal path $R^{*}\rightarrow E^{*}$ to disentangle the intertwined information can achieve higher performances on both remaining data $R_{tr}$ and $R_{ts}$. However, the forgetting performances $F_{tr}$ and $F_{ts}$ cannot be satisfying on the forgetting tasks. In comparison, the third setting where we add counterfactual samples demonstrates an obvious improvement in the forgetting performances. Meanwhile, it cannot preserve high remaining performances, especially in random data removal. Finally, CaMU, which contains all the causal paths, can reach the most similar results as the retrained model considering all the evaluation metrics.

\begin{table}[H]
\tiny
\centering
\vspace{-5mm}
\caption{Efficiency comparisons}
\label{tab:time}
\vspace{-4mm}
\setlength\tabcolsep{3pt}
\renewcommand{\arraystretch}{1.25} 
\begin{center}
\begin{tabular}{p{0.5cm}<{\centering} | p{0.9cm}<{\centering} | p{0.7cm}<{\centering}p{0.6cm}<{\centering}p{0.7cm}<{\centering}p{0.6cm}<{\centering}p{0.6cm}<{\centering}p{0.6cm}<{\centering}p{0.6cm}<{\centering}p{0.6cm}<{\centering}}  \toprule
Type &  Data &  Retrain& NGrad& {Boundary}& {T-S}& {SCRUB} & {SISA}& {Unroll}& {CaMU}\\  \midrule
 \multirow{4}{*}{\rotatebox{90}{Data}} & {Digit}& {27.38}& {19.77}& {19.58}& {37.65}& {34.61}& {44.34}& $\text{2.76}^{\star}$ & $\text{14.02}^{\dagger}$ \\
 ~& {Fashion} & {26.93}& {19.43}& {16.16}& {73.13}& {34.33}& {44.24}& $\text{2.79}^{\star}$ & $\text{14.24}^{\dagger}$ \\ 
~&{C10}& {1525.06}& $\text{73.85}^{\star}$& {579.11}& {2069.89}& {571.54}& {222.24}& $\text{48.78}^{\dagger}$ & {212.42} \\
 ~&{C100}& {1666.31}& $\text{58.90}^{\star}$& {581.56}& {2066.13}& {573.94}& {235.07}& $\text{48.78}^{\dagger}$ & {212.53} \\\midrule
  \multirow{4}{*}{\rotatebox{90}{Class}} & {Digit}& {27.29}& {18.70}& {19.49}& {73.50}& {34.49}& {45.14}& $\text{2.76}^{\star}$ & $\text{2.89}^{\dagger}$  \\
 ~& {Fashion} &  {27.01}& {13.12}& {15.89}& {75.15}& {34.54}& {43.84}& $\text{2.71}^{\star}$ & $\text{2.89}^{\dagger}$ \\ 
~&{C10}& {763.97}& {65.11}& {577.31}& {2066.89}& {573.12}& {220.34}& $\text{48.69}^{\dagger}$ & $\text{41.74}^{\star}$ \\
 ~&{C100}& {840.13}& $\text{6.92}^{\star}$& {58.48}& {2064.57}& {611.47}& {233.07}& $\text{46.10}^{\dagger}$ & {104.41} \\
\bottomrule
\end{tabular}
\vspace{-8mm}
\end{center}
\end{table}

\begin{figure}[t]
    \centering
    \vspace{-4mm}
    \subfigure[Digit]{
        \begin{minipage}[htbp]{0.44\linewidth}
            \centering
            \includegraphics[width=\textwidth,height=0.9\textwidth]{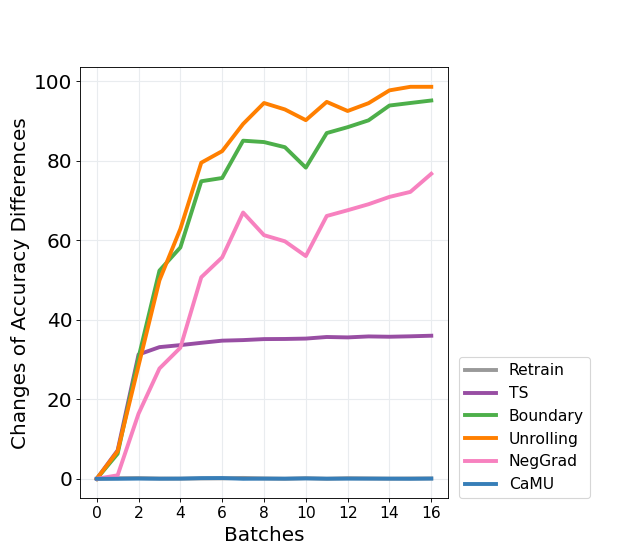}
        \end{minipage}%
    }
    \subfigure[Fashion]{
        \begin{minipage}[htbp]{0.44\linewidth}
            \centering
            \includegraphics[width=\textwidth,height=0.9\textwidth]{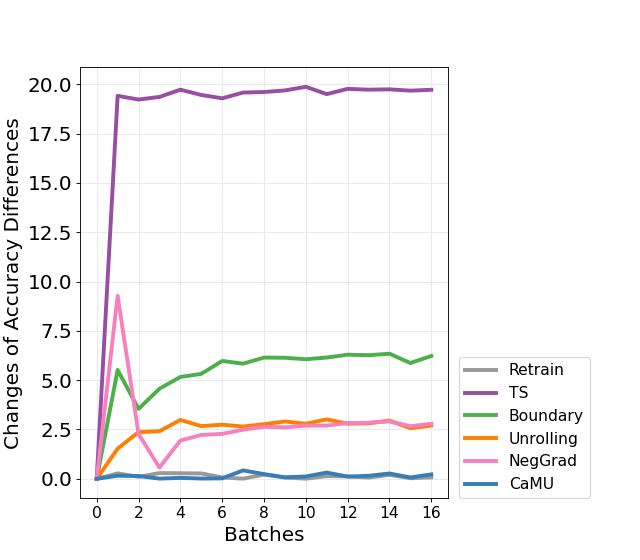}
        \end{minipage}%
    }%
    \vspace{-4mm}
    \subfigure[CIFAR10]{
        \begin{minipage}[htbp]{0.44\linewidth}
            \centering
            \vspace*{-0.4cm}
            \includegraphics[width=\textwidth,height=0.9\textwidth]{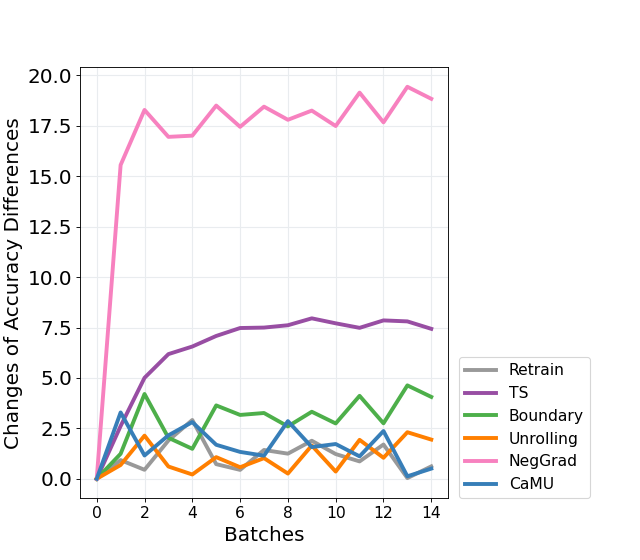}
        \end{minipage}%
    }%
    \subfigure[CIFAR100]{
        \begin{minipage}[htbp]{0.44\linewidth}
            \centering
            \vspace*{-0.4cm}
            \includegraphics[width=\textwidth,height=0.9\textwidth]{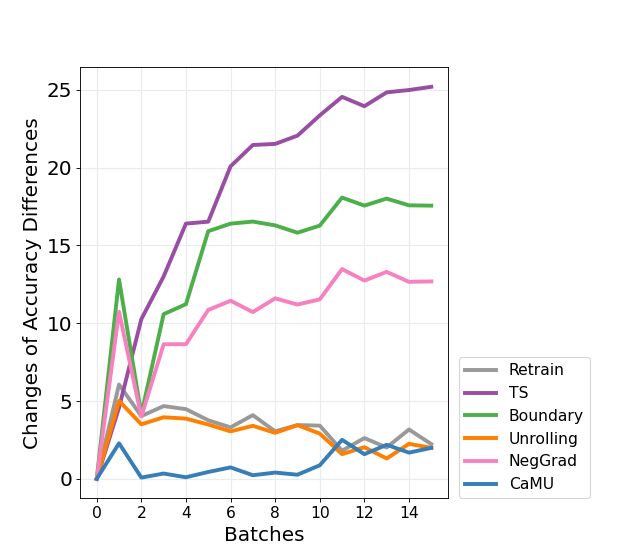}
        \end{minipage}%
    }%
    \centering
    \vspace{-1em}
    \caption{Changes of $R_{ts} - F_{ts}$ during relearning on the class removal task}
    \label{fig:compareclass}
    \vspace{-6mm}
\end{figure}

\subsection{Relearn Performances (RQ3)}

Then to examine CaMU on avoiding the residual latent information of forgetting data and performance degradation on remaining data, we finetune the post-unlearning models on the whole remaining data and the results are shown in Fig. \ref{fig:comparedata} and Fig. \ref{fig:compareclass}. By comparing the changes of differences of accuracies on remaining data and forgetting data (i.e. $R_{tr} - F_{tr}$ and $R_{ts} - F_{ts}$), we find the CaMU has the most stable performances in both random data removal and class removal tasks, which indicates that the model will not relearn new information on the forgetting or remaining data and will not continue the insufficient forgetting the by fine-tuning.

\begin{table}[H]
\tiny
\centering
\vspace{-5mm}
\caption{Ablation study on 10\% data removal task}
\label{tab:abl1}
\vspace{-4mm}
\setlength\tabcolsep{3pt}
\renewcommand{\arraystretch}{1.25} 
\begin{center}

\begin{tabular}{p{0.5cm}<{\centering} |p{0.5cm}<{\centering} p{1.0cm}<{\centering} p{1.1cm}<{\centering}p{1.2cm}<{\centering}p{1.7cm}<{\centering}p{1.0cm}<{\centering}} \toprule 
{Data} & {Metric} & {Retrain}& {Finetune}& {Remove $F^*$}& {Remove$R^*$to$E^*$} & {CaMU}\\  \midrule

 \multirow{4}{*}{\rotatebox{90}{Digit}} & $R_{tr}$& {99.59}& {98.58}& $\text{99.34}^{\star}$& {96.45}& $\text{99.24}^{\dagger}$ \\ 
                        & $F_{tr}$&  {98.79}& $\text{98.46}^{\star}$& {99.33}& {93.13}& $\text{99.12}^{\dagger}$ \\
                        & ${Ts}\uparrow$&  {98.93}& {98.40}& $\text{99.06}^{\dagger}$& {95.85}& $\text{98.98}^{\star}$ \\
                        & $\mathcal{M}$&  {49.58}& $\text{49.31}^{\star}$& $\text{49.00}^{\dagger}$& {42.42}& {47.91} \\\midrule

 \multirow{4}{*}{\rotatebox{90}{Fashion}} & $R_{tr}$&  {94.41}& $\text{93.22}^{\dagger}$& $\text{94.09}^{\star}$& {88.07}& {92.51} \\
                        & $F_{tr}$&  {91.00}& $\text{92.49}^{\dagger}$& {93.59}& {82.11}& $\text{90.96}^{\star}$ \\
                        & ${Ts}\uparrow$&  {90.51}& $\text{89.57}^{\dagger}$& $\text{90.52}^{\star}$& {83.81}& {88.69} \\
                        & $\mathcal{M}$&  {49.97}& $\text{50.33}^{\star}$& $\text{50.34}^{\dagger}$& {44.73}& {48.14} \\\midrule
 \multirow{4}{*}{\rotatebox{90}{C10}} & $R_{tr}$&  {84.25}& $\text{84.05}^{\star}$& {79.99}& {77.71}& $\text{82.19}^{\dagger}$ \\  
                        & $F_{tr}$&  {79.91}& {83.76}& $\text{79.12}^{\star}$& {76.02}& $\text{81.64}^{\dagger}$ \\
                        & ${Ts}\uparrow$&  {87.49}& $\text{88.45}^{\star}$& {85.60}& {82.82}& $\text{87.32}^{\dagger}$ \\
                        & $\mathcal{M}$& {57.96}&  $\text{55.71}^{\dagger}$& $\text{56.37}^{\star}$& {51.32}&  {54.10} \\\midrule
 \multirow{4}{*}{\rotatebox{90}{C100}} & $R_{tr}$&  {66.78}& {62.57}& $\text{63.25}^{\star}$& {61.33}& $\text{62.77}^{\dagger}$ \\ 
                        & $F_{tr}$&  {55.40}& {61.16}& {62.26}& $\text{58.89}^{\star}$& $\text{60.92}^{\dagger}$ \\
                        & ${Ts}\uparrow$&  {63.54}& {62.70}& $\text{63.47}^{\star}$& {61.38}& $\text{62.79}^{\dagger}$ \\
                        & $\mathcal{M}$&  {59.31}& $\text{56.46}^{\dagger}$& $\text{57.09}^{\star}$& {51.73}& {53.33} \\
\bottomrule
\end{tabular}
\vspace{-8mm}
\end{center}
\end{table}

\begin{table}[H]
\tiny
\centering
\vspace{-5mm}
\caption{Ablation study on class 0 removal task}
\label{tab:abl2}
\vspace{-4mm}
\setlength\tabcolsep{3pt}
\renewcommand{\arraystretch}{1.25} 
\begin{center}
\begin{tabular}{p{0.5cm}<{\centering} |p{0.5cm}<{\centering} p{1.0cm}<{\centering} p{1.1cm}<{\centering}p{1.2cm}<{\centering}p{1.7cm}<{\centering}p{1.0cm}<{\centering}} \toprule 
{Data} & {Metric} & {Retrain}& {Finetune}& {Remove $F^*$}& {Remove$R^*$to$E^*$} & {CaMU}\\ \midrule

 \multirow{3}{*}{\rotatebox{90}{Digit}} & $R_{ts}\uparrow$&  {98.81}& {99.09}& $\text{99.78}^{\star}$& {99.04}& $\text{99.22}^{\dagger}$ \\ 
                        & $F_{ts}\downarrow$&  {0}& {84.73}& {76.73}& $\text{0}^{\star}$& $\text{0}^{\star}$ \\
                        & $\mathcal{M}$&  {26.49}& $\text{27.59}^{\star}$& $\text{24.92}^{\dagger}$& {23.19}&{23.62} \\\midrule

 \multirow{3}{*}{\rotatebox{90}{Fashion}} & $R_{ts}\uparrow$& {92.66}&  $\text{92.66}^{\dagger}$& $\text{92.75}^{\star}$ & {91.78}& $\text{92.66}^{\dagger}$\\  
                        & $F_{ts}\downarrow$& {0}&  {8.36}& {1.88}& $\text{0}^{\star}$& $\text{0}^{\star}$\\
                        & $\mathcal{M}$& {38.24}&  $\text{33.77}^{\dagger}$& $\text{33.77}^{\dagger}$& {27.15}& $\text{35.24}^{\star}$  \\\midrule
 \multirow{3}{*}{\rotatebox{90}{C10}} & $R_{ts}\uparrow$& {87.01}&  {86.20}& $\text{87.04}^{\dagger}$& {86.58}& $\text{87.16}^{\star}$ \\
                        & $F_{ts}\downarrow$&  {0}& {3.30}& {4.24}& $\text{0}^{\star}$& $\text{0}^{\star}$ \\
                        & $\mathcal{M}$& {67.76}&  {60.00}& {60.13}& $\text{62.93}^{\dagger}$& $\text{63.58}^{\star}$ \\\midrule
 \multirow{3}{*}{\rotatebox{90}{C100}} & $R_{ts}\uparrow$& {61.08}&  {55.76}& $\text{56.33}^{\dagger}$& {56.13}& $\text{56.52}^{\star}$ \\ 
                        & $F_{ts}\downarrow$& {0}&  {1.4}& {1.2}& $\text{0}^{\star}$& $\text{0}^{\star}$ \\
                        & $\mathcal{M}$&  {61.96}& $\text{67.72}^{\star}$& $\text{68.49}^{\dagger}$& {71.53}& {71.25} \\
\bottomrule
\end{tabular}
\end{center}
\vspace{-6mm}
\end{table}

\section{Conclusion}

We employed causal inference techniques to tackle machine unlearning challenges, specifically focusing on the disentanglement of information between forgotten and remaining data to mitigate issues of residual latent information and performance degradation in post-unlearning models. To achieve this, we introduced causal graphs for both pre-unlearning and post-unlearning phases to analyze the causes of the latent information and performance degradation issues. Based on the analysis, we reframe deep model unlearning as a causal effect removal problem and introduced the Causal Machine Unlearning (CaMU) framework. Within CaMU, we eliminated undesirable causal effects using counterfactual data for forgetting samples while maintaining the essential information by preserving the causal effects associated with the remaining data. To validate the effectiveness and efficiency of CaMU, we undertook extensive experiments on four real-world datasets and provide insightful analysis. The experiments showed significant improvements in CaMU compared with other works. Additionally, ablation studies substantiated the accuracy of the proposed causal graphs, thereby fortifying the robustness and credibility of our approach.

\bibliographystyle{siam}
\bibliography{arxiv}

\newpage
\renewcommand\thesection{\Alph{section}}
\setcounter{section}{0}
\section{Supplementary Material}
\subsection{Related Works}
In this section, we will introduce some related works on Machine unlearning and the application of {causal analysis} in deep learning
\subsubsection{Machine Unlearning}
Currently, two types of approaches to machine unlearning are employed. The first type is \textit{exact unlearning}, which requires the unlearned model to achieve the same level of performance as the retrained model, in terms of both model parameters and prediction accuracy. Exact unlearning is commonly applied to classical machine learning models with simple structures and analytical optimization solutions \cite{RF,sisa,Two-stageModelRetraining}. The second type is \textit{approximate unlearning}, which requires the unlearned model to get similar performances to the retrained model on both the remaining data and the target data to be forgotten. Approximate unlearning methods are widely applied to deep models  \cite{Variational, fastyet, DBLP:conf/cvpr/MehtaPSR22, Sunshine,unroll,Amnesiac,boundary,scrub,ts}. For example, the deep model approximate unlearning consider generating a series of noisy data for unlearning tasks \cite{fastyet}, or recovering the changes of parameters occurring in the training of data to be forgotten \cite{unroll}.
\subsubsection{Causal Analysis}
The causal inference techniques, especially {causal analysis}, have been widely used in deep learning research. The first type is the {causal discovery which can assist in constructing the causal structures inside the data and models. The causal structure can show the underlying data-generating process across different environments, which can be explored based on some assumptions from the graphical or structural causal models. Currently, causal discovery is a widely used technique in solving domain adaptation problems. For example, \cite{DBLP:conf/aaai/ZhangGS15} shows a systematic view of applying causality to the multi-source domain adaptation while \cite{DBLP:journals/corr/abs-1804-04333} propose the CG-DAN to learn the causal structure with latent variables. Another type of work refers to the application of causal inference in deep learning problems. Specifically, the causal inference techniques have been widely used in fairness deep learning \cite{DBLP:conf/acl/0003FWMX20, DBLP:conf/recsys/LiuCZDHP021, DBLP:conf/kdd/WeiFCWYH21}.  These works propose causal graphs for the specific problem based on the previous experimental phenomena and then add intervention on the relationships between the variables in causal graphs. For example, \cite{DBLP:conf/acl/0003FWMX20} employs counterfactual approach to eliminate the causal effect of the potential source of bias in text classification. \cite{DBLP:conf/recsys/LiuCZDHP021} and \cite{DBLP:conf/kdd/WeiFCWYH21} applied backdoor adjustments and counterfactual approach to remove the causal effects of the bias in the recommender system.
}

\subsection{Implementation Details}

\subsubsection{Datasets and Models}

To validate the effectiveness of the CaMU framework, we conduct experiments on four different datasets: \textbf{Digit} \cite{mnist}, \textbf{Fashion} \cite{fashion}, \textbf{C10} \cite{cifar} and \textbf{C100} \cite{cifar}. Among them, Digit and Fashion are both MNIST datasets. Detailedly, Digit consists of images of handwritten digits from 0 to 9 and Fashion contains ten common clothes. Both MNIST datasets contain 60,000 training samples and 10,000 test samples. The data samples in the two MNIST datasets are all 28 × 28 grayscale images. C10 (CIFAR-10) and C100 (CIFAR-100) contain images of ten and one hundred, respectively, common animals or items in our lives. Both CIFAR datasets contain 50,000 training samples and 10,000 test samples each of which is an RGB image in the shape of 32 × 32.

For the two MNIST datasets, we use a convolutional neural network (\textbf{CNN}) with two convolutional layers \cite{cnn} and three linear layers while for the two CIFAR datasets, we choose an \textbf{18-layer ResNet} backbone\cite{resnet} without pre-trained parameters. To get pre-unlearning models, we train the CNN models for 10 epochs on two MNIST datasets to get stable training and test loss and we train the 18-layer ResNet models for 20 epochs on two CIFAR datasets. In the following experiments on CaMU and other baselines, we use the same pre-unlearning model for the unlearning tasks.

\subsubsection{Baselines}
We compare the performance of \textbf{CaMU} with seven baseline results including \textbf{Retrain} which are the golden standards of unlearning, and six state-of-the-art unlearning works on deep models: as described following. \textbf{NGrad} finetunes the pre-unlearning model with positive gradients on remaining data and negative gradients on forgetting data. \textbf{Boundary} \cite{boundary} shrinks the distance between the decision boundaries of the forgetting data and remaining data to eliminate differences in predictions. \textbf{T-S} \cite{ts} proposes to retrain two teacher models on forgetting data and remaining data and then use the output of the teacher model to instruct the forgetting of the pre-unlearning model. \textbf{SCRUB} \cite{scrub} also uses a teacher-student framework to instruct the model to be consistent with the pre-unlearning model on remaining data and inconsistent with the pre-unlearning model on forgetting data. \textbf{SISA}~\cite{sisa} proposes a distributed approach for unlearning that retrains the small data shards from the remaining dataset in different models and ensemble the final results with less time consumption. \textbf{Unroll}~\cite{unroll} performs incremental training with the forgotten data in the first batch. It records gradients when learning the first batch and adds recorded gradients on weights after the incremental training.

\subsubsection{Environment} 
All the experiments are conducted on one server NVIDIA RTX A6000 GPU (48GB GDDR6 Memory) and 12th Gen Intel(R) Core(TM) i9-12900K (16 cores and 128GB Memory). The code of CaMU was implemented in Python 3.9.16 and Cuda 11.6.1. The main Python packages' versions are the following: Numpy 1.23.5; Pandas 2.0.1; Pytorch 1.13.1; Torchvision 0.14.1.  The datasets in experiments: \textbf{Digit-MNIST} \cite{mnist}, \textbf{Fashion-MNIST} \cite{fashion}, \textbf{C10} \cite{cifar}, and \textbf{C100} dataset \cite{cifar} are all downloaded from the Torchvision library. Moreover, all the comparison methods provide open resources for their implementation code: \textbf{Boundary} \footnote{https://www.dropbox.com/s/bwu543qsdy4s32i/Boundary-Unlearning-Code.zip?dl=0}, \textbf{T-S} \footnote{ https://github.com/vikram2000b/bad-teaching-unlearning}, \textbf{SCRUB} \footnote{ https://github.com/meghdadk/SCRUB}, \textbf{SISA} \footnote{ https://github.com/cleverhans-lab/machine-unlearning}, \textbf{Unroll} \footnote{https://github.com/cleverhans-lab/unrolling-sgd}.
\subsubsection{Initializations} 
For the experiment models, we choose the a \textbf{CNN}\cite{cnn} with two convolutional layers for the two MNIST datasets. The output channels for the two convolutional layers are 16 and 32 respectively. Then the other parts of the CNN consist of three linear layers with the output dimensions as 256, 128 and 10. For the two CIFAR datasets, we choose an \textbf{18-layer ResNet} \cite{resnet} and without the pre-trained weights. The hyperparameters for each model and dataset are recorded on the code page \footnote{https://github.com/ShaofeiShen768/CaMU}.

All the experiments are based on the original models trained in the four datasets. We train two CNN models on two MNIST datasets for 10 epochs with a learning rate of 0.001 while we train another two 18-layer ResNet models on two CIFAR datasets for 20 epochs with a learning rate of 0.00005. For the golden standard baselines \textbf{Retrain}, we retrain the CNN models on two MNIST datasets for 20 epochs with a learning rate of 0.001. We retrain the 18-layer ResNet models on two CIFAR datasets for 40 epochs with a learning rate of 0.00005. Then for the other six comparison baselines:\textbf{NegGrad}, \textbf{Boundary}\cite{boundary}, \textbf{T-S}\cite{ts}, \textbf{SCRUB}\cite{scrub}, \textbf{SISA}\cite{sisa}, \textbf{Unroll}\cite{unroll}, we keep the training and unlearning module as the source code and rewrite the parts of the data splitting to change the settings of the data to be forgotten. We keep the hyperparameters of the training process the same as in this paper and adjust other necessary parameters for the unlearning stage to get as high performances as we can. 

\begin{table}[H]
\tiny
\centering
\caption{Performances of 15\% data removal task (\%). $\star$ means the best results and $\dagger$ stands for the second best. The two notations have the same meanings in the following tables.}
\label{tab:randomsup1}
\vspace{-3mm}
\setlength\tabcolsep{3pt}
\renewcommand{\arraystretch}{1.25} 
\begin{center}
\begin{tabular}{p{0.5cm}<{\centering} |p{0.5cm}<{\centering} p{0.7cm}<{\centering}p{0.7cm}<{\centering}p{0.7cm}<{\centering}p{0.7cm}<{\centering}p{0.7cm}<{\centering}p{0.7cm}<{\centering}p{0.7cm}<{\centering}p{0.7cm}<{\centering}}  \toprule
Data & Metric & Retrain& NGrad& {Boundary}& {T-S}& {SCRUB} & {SISA}& {Unroll}& {CaMU}\\  \midrule

 \multirow{3}{*}{\rotatebox{90}{Digit}} & $R_{tr}$& {99.56}& {99.18}& {97.15}& $\text{99.34}^{\star}$& $\text{99.27}^{\dagger}$& {87.98}& {99.74} & {99.10} \\ 
                        & $F_{tr}$& {98.84}& $\text{98.86}^{\dagger}$& {97.13}& {96.50}& {99.24}& {88.24}& {99.66} & $\text{98.83}^{\star}$ \\ 
                        & $Ts\uparrow$& {99.04}& {98.62}& {96.62}& {98.02}& $\text{98.98}^{\star}$& {87.77}& $\text{99.18}^{\dagger}$ & {98.76} \\ \midrule
 \multirow{3}{*}{\rotatebox{90}{Fashion}} & $R_{tr}$& {94.86}& {93.26}& {48.94}& {92.63}& {91.19}& {82.56}& $\text{93.67}^{\star}$ & $\text{92.83}^{\dagger}$ \\  
                        & $F_{tr}$& {91.02}& {92.38}& {48.39}& {88.13}& $\text{90.91}^{\dagger}$& {82.40}& {93.34} & $\text{91.12}^{\star}$  \\ 
                        & $Ts\uparrow$& {90.56}& {89.32}& {47.58}& {88.85}& $\text{89.33}^{\dagger}$& {80.61}& $\text{89.61}^{\star}$  & {89.17} \\ \midrule
 \multirow{3}{*}{\rotatebox{90}{C10}} & $R_{tr}$& {79.08}& {72.41}& {52.63}& $\text{78.03}^{\dagger}$& {27.43}& {74.47}& {77.65} & $\text{78.97}^{\star}$ \\  
                        & $F_{tr}$& {76.30}& {70.71}& {52.95}& $\text{75.44}^{\dagger}$& {27.61}& {74.67}& $\text{77.09}^{\star}$ & {78.31} \\ 
                        & $Ts\uparrow$& {84.07}& {79.99}& {62.09}& {85.40}& {29.60}& {68.12}& $\text{83.73}^{\star}$ & $\text{84.52}^{\dagger}$ \\ \midrule
 \multirow{3}{*}{\rotatebox{90}{C100}} & $R_{tr}$& {58.39}& {48.28}& {32.69}& {54.40}& {8.27}& $\text{55.65}^{\dagger}$& {50.40} & $\text{59.86}^{\star}$ \\ 
                        & $F_{tr}$& {50.96}& {43.39}& {32.55}& $\text{48.83}^{\star}$& {8.36}& $\text{55.71}^{\dagger}$& {49.26} & {59.02} \\
                        & $Ts\uparrow$& {59.44}& {50.84}& {38.23}& $\text{57.45}^{\dagger}$& {8.65}& {38.53}& {54.11} & $\text{61.68}^{\star}$ \\
\bottomrule
\end{tabular}
\end{center}
\vspace{-5mm}
\end{table}

\begin{table}[H]
\tiny
\centering
\vspace{-3mm}
\caption{Performances of 20\% data removal task (\%).}
\label{tab:randomsup2}
\vspace{-3mm}
\setlength\tabcolsep{3pt}
\renewcommand{\arraystretch}{1.25} 
\begin{center}
\begin{tabular}{p{0.5cm}<{\centering} |p{0.5cm}<{\centering} p{0.7cm}<{\centering}p{0.7cm}<{\centering}p{0.7cm}<{\centering}p{0.7cm}<{\centering}p{0.7cm}<{\centering}p{0.7cm}<{\centering}p{0.7cm}<{\centering}p{0.7cm}<{\centering}}  \toprule
Data & Metric & Retrain& NGrad& {Boundary}& {T-S}& {SCRUB} & {SISA}& {Unroll}& {CaMU}\\  \midrule

 \multirow{3}{*}{\rotatebox{90}{Digit}} & $R_{tr}$& {99.59}& {96.22}& {97.15}& $\text{99.32}^{\dagger}$& {99.28}& {88.02}& $\text{99.66}^{\star}$ & {99.05} \\ 
                        & $F_{tr}$& {98.71}& {95.72}& {97.13}& {96.82}& $\text{99.23}^{\dagger}$ & {88.11}& {99.58} & $\text{98.79}^{\star}$ \\ 
                        & $Ts\uparrow$& {98.84}& {95.64}& {96.62}& {98.01}& $\text{99.02}^{\dagger}$ & {87.78}& {99.12} & $\text{98.70}^{\star}$ \\ \midrule
 \multirow{3}{*}{\rotatebox{90}{Fashion}} & $R_{tr}$& {96.66}& {84.52}& {48.94}& {92.64}& {91.24}& {82.60}& $\text{94.46}^{\star}$ & $\text{93.09}^{\dagger}$ \\  
                        & $F_{tr}$& {90.58}& {83.45}& {48.39}& {88.28}& $\text{91.06}^{\star}$& {82.45}& {94.18} & $\text{91.54}^{\dagger}$ \\ 
                        & $Ts\uparrow$& {90.10}& {81.80}& {47.58}& {89.02}& {89.45}& {80.76}& $\text{90.43}^{\star}$ & $\text{89.58}^{\dagger}$ \\ \midrule
 \multirow{3}{*}{\rotatebox{90}{C10}} & $R_{tr}$& {83.44}& {43.98}& {52.63}& $\text{78.61}^{\dagger}$& {28.57}& {72.62}& {74.81} & $\text{81.58}^{\star}$ \\  
                        & $F_{tr}$& {79.10}& {40.94}& {52.95}& $\text{75.71}^{\dagger}$& {28.27}& {72.58}& {74.27} & $\text{80.52}^{\star}$ \\ 
                        & $Ts\uparrow$& {86.34}& {49.93}& {62.09}& $\text{85.42}^{\dagger}$& {31.23}& {65.49}& {80.63} & $\text{86.94}^{\star}$ \\ \midrule
 \multirow{3}{*}{\rotatebox{90}{C100}} & $R_{tr}$& {66.22}& {23.21}& {32.69}& {53.71}& {9.03}& $\text{55.51}^{\dagger}$& {49.95} & $\text{64.64}^{\star}$ \\ 
                        & $F_{tr}$& {53.72}& {18.64}& {32.55}& $\text{48.87}^{\dagger}$& {8.93}& $\text{55.59}^{\star}$& {48.57} & {63.22} \\
                        & $Ts\uparrow$& {62.01}& {26.03}& {38.23}& $\text{56.61}^{\dagger}$& {9.23}& {38.19}& {52.75} & $\text{64.92}^{\star}$ \\
\bottomrule
\end{tabular}

\end{center}
\vspace{-5mm}
\end{table}

\begin{table}[H]
\tiny
\centering
\vspace{-3mm}
\caption{Performances of class 1 removal task (\%)}
\label{tab:classsup1}
\vspace{-4mm}
\setlength\tabcolsep{3pt}
\renewcommand{\arraystretch}{1.25} 
\begin{center}
\begin{tabular}{p{0.5cm}<{\centering} |p{0.5cm}<{\centering} p{0.7cm}<{\centering}p{0.7cm}<{\centering}p{0.7cm}<{\centering}p{0.7cm}<{\centering}p{0.7cm}<{\centering}p{0.7cm}<{\centering}p{0.7cm}<{\centering}p{0.7cm}<{\centering}}  \toprule
Data & Metric & Retrain& NGrad& {Boundary}& {T-S}& {SCRUB} & {SISA}& {Unroll}& {CaMU}\\ \midrule

 \multirow{2}{*}{\rotatebox{90}{Digit}} & $R_{ts}\uparrow$& {99.07}& {13.34}& {98.81}& {98.97}& {99.01}& {99.00}& $\text{99.07}^{\dagger}$ & $\text{99.09}^{\star}$ \\
                        & $F_{ts}\downarrow$& {0}& $\text{0}^{\star}$& {95.79}& {2.87}& {96.63}& $\text{0}^{\star}$& {95.98} & $\text{0}^{\star}$ \\\midrule

  \multirow{2}{*}{\rotatebox{90}{Fash-}\rotatebox{90}{ion}} & $R_{ts}\uparrow$ & {89.59}& {11.37}& {83.59}& {89.74}& {88.44}& {89.52}& $\text{90.00}^{\star}$ & $\text{89.75}^{\dagger}$  \\
                        & $F_{ts}\downarrow$&  {0}& $\text{0}^{\star}$& {32.84}& {4.20}& {27.68}& $\text{0}^{\star}$& {94.17} & $\text{0}^{\star}$ \\\midrule
 \multirow{2}{*}{\rotatebox{90}{C10}} & $R_{ts}\uparrow$& {87.73}& {38.16}& {81.65}& $\text{84.86}^{\dagger}$ & {29.65}& {72.80}& {76.70} & $\text{85.83}^{\star}$ \\ 
                        & $F_{ts}\downarrow$&   {0}& $\text{0}^{\star}$& {0.98}& {18.64}& $\text{0}^{\star}$& $\text{0}^{\star}$& $\text{0}^{\star}$ & $\text{0}^{\star}$ \\\midrule
 \multirow{2}{*}{\rotatebox{90}{C100}} & $R_{ts}\uparrow$& {64.71}& {54.73}& {53.99}& $\text{59.45}^{\dagger}$ & {8.36}& {38.50}& {61.39} & $\text{61.90}^{\star}$ \\
                        & $F_{ts}\downarrow$& {0}& {1.40}& {2.20}& {13.20}& $\text{0}^{\star}$& $\text{0}^{\star}$& $\text{0}^{\star}$ & $\text{0}^{\star}$ \\
\bottomrule
\end{tabular}

\end{center}
\vspace{-5mm}
\end{table}

\begin{table}[H]
\tiny
\centering
\vspace{-3mm}
\caption{Performances of class 2 removal task (\%)}
\label{tab:classsup2}
\vspace{-4mm}
\setlength\tabcolsep{3pt}
\renewcommand{\arraystretch}{1.25} 
\begin{center}
\begin{tabular}{p{0.5cm}<{\centering} |p{0.5cm}<{\centering} p{0.7cm}<{\centering}p{0.7cm}<{\centering}p{0.7cm}<{\centering}p{0.7cm}<{\centering}p{0.7cm}<{\centering}p{0.7cm}<{\centering}p{0.7cm}<{\centering}p{0.7cm}<{\centering}}  \toprule
Data & Metric & Retrain& NGrad& {Boundary}& {T-S}& {SCRUB} & {SISA}& {Unroll}& {CaMU}\\ \midrule

 \multirow{2}{*}{\rotatebox{90}{Digit}} & $R_{ts}\uparrow$& {99.10}& {97.18}& {98.72}& {98.89}& {99.09}& {99.07}& $\text{99.25}^{\star}$ & $\text{99.14}^{\dagger}$ \\
                        & $F_{ts}\downarrow$& {0}& {46.72}& {88.95}& {4.86}& {88.72}& $\text{0}^{\star}$& {99.02} & $\text{0}^{\star}$ \\\midrule

  \multirow{2}{*}{\rotatebox{90}{Fash-}\rotatebox{90}{ion}} & $R_{ts}\uparrow$ & {92.50}& {86.28}& {87.46}& {92.12}& {90.94}& $\text{92.17}^{\dagger}$& {88.72} & $\text{92.46}^{\star}$ \\
                        & $F_{ts}\downarrow$&  {0}& $\text{0}^{\star}$& {1.72}& {22.62}& $\text{0}^{\star}$& $\text{0}^{\star}$& {0.40} & $\text{0}^{\star}$ \\\midrule
 \multirow{2}{*}{\rotatebox{90}{C10}} & $R_{ts}\uparrow$& {86.75}& {55.60}& {83.09}& $\text{87.26}^{\dagger}$& {32.33}& {72.04}& {84.30} & $\text{87.61}^{\star}$ \\ 
                        & $F_{ts}\downarrow$&   {0}& $\text{0}^{\star}$& {2.42}& {9.76}& $\text{0}^{\star}$& $\text{0}^{\star}$& $\text{0}^{\star}$ & $\text{0}^{\star}$ \\\midrule
 \multirow{2}{*}{\rotatebox{90}{C100}} & $R_{ts}\uparrow$& {61.42}& {58.92}& {54.28}& {59.80}& {7.91}& {38.92}& $\text{60.31}^{\dagger}$ & $\text{62.05}^{\star}$ \\
                        & $F_{ts}\downarrow$& {0}& {10.80}& {3.20}& {22.00}& $\text{0}^{\star}$& $\text{0}^{\star}$& $\text{0}^{\star}$ & $\text{0}^{\star}$ \\
\bottomrule
\end{tabular}
\end{center}
\vspace{-5mm}
\end{table}

\begin{table}[H]
\tiny
\centering
\vspace{-5mm}
\caption{Attack success rate comparisons in MIA}
\label{tab:miasup1}
\vspace{-3mm}
\setlength\tabcolsep{3pt}
\renewcommand{\arraystretch}{1.25} 
\begin{center}
\begin{tabular}{p{0.5cm}<{\centering} | p{0.8cm}<{\centering} | p{0.7cm}<{\centering}p{0.6cm}<{\centering}p{0.7cm}<{\centering}p{0.8cm}<{\centering}p{0.6cm}<{\centering}p{0.6cm}<{\centering}p{0.6cm}<{\centering}p{0.6cm}<{\centering}}  \toprule
Type &  Data &  Retrain& NGrad& {Boundary}& {T-S}& {SCRUB} & {SISA}& {Unroll}& {CaMU}\\  \midrule
 \multirow{4}{*}{\rotatebox{90}{15\% Data}} & {Digit}& {49.79}& {50.24}& {49.97}& {40.78}& {49.71}& {40.00}& {49.43} & {48.93} \\
 ~& {Fashion} & {50.33}& {50.06}& {50.02}& {44.56}& {50.14}& {40.00}& {49.63} & {50.16} \\
~&{C10}& {57.31}& {56.23}& {59.57}& {55.31}& {54.92}& {44.44}& {57.60} & {54.97} \\
 ~&{C100}&{58.39}& {56.69}& {59.69}& {56.50}& {54.91}& {44.44}& {58.71} & {55.44} \\\midrule
  \multirow{4}{*}{\rotatebox{90}{20\% Data}} & {Digit}& {50.05}& {50.34}& {50.38}& {41.25}& {50.35}& {33.33}& {50.56} & {49.12} \\
 ~& {Fashion} & {49.92}& {50.42}& {50.20}& {43.47}& {49.91}& {33.33}& {50.52} & {50.24} \\
~&{C10}& {56.82}& {53.79}& {59.40}& {56.30}& {54.99}& {37.50}& {57.42} & {53.49} \\
 ~&{C100}&{57.86}& {53.21}& {59.35}& {56.47}& {55.03}& {37.50}& {57.37} & {54.87} \\\midrule
  \multirow{4}{*}{\rotatebox{90}{Class 1}} & {Digit}& {23.42}& {52.57}& {27.24}& {21.86}& {24.91}& {47.09}& {26.24} & {22.11} \\
 ~& {Fashion} &  {33.70}& {46.47}& {32.06}& {24.39}& {28.28}& {50.00}& {35.76} & {24.88} \\
~&{C10}& {63.35}& {61.10}& {59.70}& {41.63}& {45.24}& {50.12}& {58.04} & {66.18} \\
 ~&{C100}& {64.60}& {61.28}& {74.62}& {76.80}& {53.28}& {50.59}& {70.18} & {71.78} \\\midrule
   \multirow{4}{*}{\rotatebox{90}{Class 2}} & {Digit}& {28.65}& {37.56}& {38.12}& {23.03}& {34.98}& {50.11}& {40.24} & {23.51} \\
 ~& {Fashion} &  {40.39}& {34.63}& {41.27}& {25.23}& {34.20}& {50.00}& {40.61} & {28.08} \\
~&{C10}& {69.58}& {53.15}& {65.46}& {45.25}& {58.55}& {50.12}& {67.59} & {65.15} \\
 ~&{C100}& {56.07}& {62.57}& {71.76}& {74.39}& {51.70}& {50.59}& {64.03} & {62.19} \\\midrule
 \multicolumn{2}{c}{Average Diff $\downarrow$}& {-}& {1.25}& {2.41}& {5.19}& {3.70}& {6.32}& {1.48} & {1.82} \\
\bottomrule
\end{tabular}
\end{center}
\vspace{-6mm}
\end{table}

\begin{table}[H]
\tiny
\centering
\caption{Algorithm Efficiency}
\label{tab:timesup1}
\vspace{-3mm}
\setlength\tabcolsep{3pt}
\renewcommand{\arraystretch}{1.25} 
\begin{center}
\begin{tabular}{p{0.5cm}<{\centering} | p{0.8cm}<{\centering} | p{0.7cm}<{\centering}p{0.6cm}<{\centering}p{0.7cm}<{\centering}p{0.7cm}<{\centering}p{0.6cm}<{\centering}p{0.7cm}<{\centering}p{0.6cm}<{\centering}p{0.6cm}<{\centering}}  \toprule
Type &  Data &  Retrain& NGrad& {Boundary}& {T-S}& {SCRUB} & {SISA}& {Unroll}& {CaMU}\\  \midrule
 \multirow{4}{*}{\rotatebox{90}{15\% Data}} & {Digit}& {26.21}& {9.42}& {23.70}& {98.32}& {42.98}& {10.60}& {2.98} & {18.87} \\
 ~& {Fashion} & {25.99}& {9.52}& {23.73}& {98.61}& {43.23}& {10.58}& {3.03} & {18.84} \\
~&{C10}& {830.53}& {72.23}& {569.15}& {2110.50}& {576.40}& {72.28}& {52.66} & {163.31} \\
 ~&{C100}&{831.73}& {72.14}& {569.38}& {2100.38}& {575.88}& {78.26}& {52.32} & {163.96} \\\midrule
  \multirow{4}{*}{\rotatebox{90}{20\% Data}} & {Digit}& {36.40}& {18.05}& {17.25}& {74.51}& {32.39}& {8.84}& {3.02} & {22.55} \\
 ~& {Fashion} & {36.19}& {17.92}& {17.23}& {75.15}& {32.14}& {8.81}& {3.00} & {22.30} \\
~&{C10}& {1483.85}& {912.47}& {731.16}& {2073.63}& {1242.01}& {44.14}& {52.74} & {173.30} \\
 ~&{C100}&{1490.37}& {466.16}& {730.94}& {2137.80}& {1684.85}& {46.85}& {52.54} & {216.53} \\\midrule
  \multirow{4}{*}{\rotatebox{90}{Class 1}} & {Digit}& {40.06}& {11.96}& {10.07}& {74.93}& {34.19}& {9.09}& {2.82} & {3.01} \\
 ~& {Fashion} &  {40.66}& {11.91}& {9.06}& {73.95}& {34.43}& {8.90}& {2.80} & {2.75} \\
~&{C10}& {1667.71}& {92.87}& {588.17}& {2070.08}& {570.98}& {45.70}& {48.49} & {43.26} \\
 ~&{C100}& {1837.43}& {11.23}& {36.63}& {2068.82}& {611.87}& {45.03}& {46.37} & {109.15} \\\midrule
   \multirow{4}{*}{\rotatebox{90}{Class 2}} & {Digit}& {27.65}& {3.54}& {15.99}& {98.28}& {45.20}& {10.80}& {2.83} & {2.96} \\
 ~& {Fashion} &  {27.60}& {3.64}& {16.30}& {99.96}& {44.80}& {11.02}& {2.86} & {2.97} \\
~&{C10}& {880.64}& {48.02}& {379.10}& {2101.11}& {597.19}& {72.04}& {47.70} & {43.29} \\
 ~&{C100}& {967.05}& {5.87}& {38.49}& {2100.15}& {639.26}& {77.43}& {46.33} & {108.85} \\
\bottomrule
\end{tabular}

\end{center}
\end{table}

\subsection{Additional Experiments}

The following table \ref{tab:randomsup1}, \ref{tab:randomsup2}, \ref{tab:classsup1}, \ref{tab:classsup2}, \ref{tab:miasup1}, and \ref{tab:timesup1} present the results of additional experiments. All the experiment results are the average value under 5 different random trials. Firstly, table \ref{tab:random1} and \ref{tab:randomsup2} demonstrate the additional results on the 15\% data removal and 20\% data removal tasks. In the results of table \ref{tab:randomsup1}, CaMU can reach relatively higher performance compared with other methods. However, CaMU only reaches the third-best performance under the test accuracy in the two MNIST datasets. In the result of the table \ref{tab:randomsup2}, CaMU can achieve nearly all the best performances in the forgetting data and test data. Secondly, the following two tables \ref{tab:classsup1} and \ref{tab:classsup2} show the results of class 1 and class 2 removal on the four datasets. The experiment results are also consistent with the results in the main paper, where almost all the results can reach the highest performances, apart from two results that rank second. The third part of the experiments is the MIA comparison, which is shown in table \ref{tab:miasup1}. The average differences of MIA compared with the retrained models indicate the average efficacy of algorithms w.r.t MIA. The proposed CaMU can reach the third highest performance. Last but not least, table \ref{tab:timesup1} presents the time costs in the above experiments. The results are consistent with the results in the main paper as well. The cost of CaMU relies on the size of the forgetting dataset. Therefore, the time cost of CaMU on the 15\% and 20\% data removal tasks is higher than the 10\% data removal tasks. However, it can still achieve the fourth-highest efficiency in all the seven algorithms. In the two class removal tasks, CaMu can still reach the highest efficiency on two MNIST datasets and CIFAR10 datasets while the time costs on CIFAR100 are a bit higher because we repeatedly select forgetting data during unlearning. 

\end{document}